\documentclass{article}

     \usepackage[final]{neurips_2024}

\usepackage[utf8]{inputenc} %
\usepackage[T1]{fontenc}    %
\usepackage{url}            %
\usepackage{booktabs,wrapfig}       %
\usepackage{amsfonts}       %
\usepackage{nicefrac}       %
\usepackage{microtype}      %
\usepackage{xcolor}         %

\usepackage{parskip}

\usepackage{natbib}
\usepackage{amsfonts,bm}

\usepackage{amsmath, amssymb,mathrsfs, amsthm, mathtools}
\usepackage{xspace}
\usepackage{enumitem}
\usepackage{lipsum}
\usepackage{xpatch}
\usepackage{mathtools}
\usepackage{dsfont}
\usepackage{pgf,tikz}
\usetikzlibrary{positioning,matrix,patterns}
\usepackage{caption}
\usepackage{graphicx}
\usepackage{xcolor}
\usepackage{makecell}
\usepackage{multirow}
\usepackage[mathscr]{euscript}
\definecolor{hanblue}{rgb}{0.27, 0.42, 0.81}
\definecolor{deepred}{HTML}{900C3F}
\definecolor{deepgreen}{HTML}{2F6960}
\usepackage[hidelinks]{hyperref}
\hypersetup{
    colorlinks=true,
    linkcolor=hanblue,
    urlcolor=hanblue,
    citecolor=hanblue,
    anchorcolor=hanblue}

 \usepackage[capitalize,noabbrev]{cleveref}

\theoremstyle{plain}
\newtheorem{theorem}{Theorem}
\newtheorem{proposition}[theorem]{Proposition}

\newtheorem{lemma}[theorem]{Lemma}

\newtheorem{mydef}[theorem]{Definition}
\theoremstyle{definition}

\theoremstyle{remark}

\def\vx{{\mathbf{x}}}
\def\vy{{\mathbf{y}}}

\def\mP{{\bm{P}}}

\def\mX{{\mathbf{X}}}
\def\mY{{\mathbf{Y}}}

\def\calF{{\mathcal{F}}}

\def\calP{{\mathcal{P}}}

\def\sR{{\mathbb{R}}}

\newcommand{\ith}[1]{^{(#1)}}

\newcommand{\WD}{W}

\newcommand{\Exp}{\mathbb{E}}

\newcommand{\diag}{\mathrm{diag}}
\DeclareMathOperator*{\argmax}{arg\,max}

\DeclareMathOperator*{\argmin}{arg\,min}

\DeclarePairedDelimiter\abs{\lvert}{\rvert}%
\DeclarePairedDelimiter\norm{\lVert}{\rVert}%

\makeatletter
\let\oldabs\abs
\def\abs{\@ifstar{\oldabs}{\oldabs*}}
\let\oldnorm\norm
\def\norm{\@ifstar{\oldnorm}{\oldnorm*}}
\makeatother

\DeclareMathOperator{\error}{error}

\def\algo{{\textsc{ProgOT}}\xspace}

\newcommand{\eps}{\varepsilon}
\newcommand{\dd}{\mathrm{d}}

\def\E{{\mathbb{E}}}

\def\R{{\mathbb{R}}}

\newcommand*{\defeq}{\coloneqq}

\newcommand{\Tprog}{\mathscr{T}_{\mathrm{Prog}}}
\newcommand{\Tent}{\hat{T}_{\eps}}
\newcommand{\myTent}{\mathscr{E}}
\newcommand{\myS}{\mathscr{S}}

\newcommand{\muent}{\hat{\mu}_\eps}

\newcommand{\bff}{\mathbf{f}}
\newcommand{\ba}{\mathbf{a}}
\newcommand{\bx}{\mathbf{x}}
\newcommand{\bX}{\mathbf{X}}
\newcommand{\bY}{\mathbf{Y}}
\newcommand{\bg}{\mathbf{g}}
\newcommand{\bb}{\mathbf{b}}
\newcommand{\by}{\mathbf{y}}

\DeclareMathOperator{\mymin}{min}
\DeclareMathOperator{\Id}{Id}
\DeclareMathOperator{\mine}{\mymin_\varepsilon}

\usepackage{cleveref,wrapfig}
\usepackage{algpseudocode}
\usepackage{algorithm}

\allowdisplaybreaks %
\title{Progressive Entropic Optimal Transport Solvers}

\author{%
  Parnian Kassraie\\
  ETH Zurich, Apple\\
  \texttt{pkassraie@ethz.ch} \\
  \And
  Aram-Alexandre Pooladian\\
  New York University\\
  \texttt{aram-alexandre.pooladian@nyu.edu} \\
  \And
  Michal Klein\\
  Apple\\
  \texttt{michalk@apple.com} \\
  \AND
  James Thornton\\
  Apple\\
  \texttt{jamesthornton@apple.com} \\
  \And
  Jonathan Niles-Weed\\
  New York University\\
  \texttt{jnw@cims.nyu.edu} \\
  \And
  Marco Cuturi\\
  Apple\\
  \texttt{cuturi@apple.com} \\
}

\begin{document}

\maketitle
\begin{abstract}
Optimal transport (OT) has profoundly impacted machine learning 
by providing theoretical and computational tools to realign datasets.
In this context, given two large point clouds of sizes $n$ and $m$ in $\mathbb{R}^d$, entropic OT (EOT) solvers have emerged as the most reliable tool to either solve the \citeauthor{Kan42} problem and output a $n\times m$ coupling matrix, or to solve the \citeauthor{Monge1781} problem and learn a vector-valued push-forward map. 
While the robustness of EOT couplings/maps makes them a go-to choice in practical applications, EOT solvers remain difficult to tune because of a small but influential set of hyperparameters, notably the omnipresent entropic regularization strength $\varepsilon$. Setting $\varepsilon$ can be difficult, as it simultaneously impacts various performance metrics, such as compute speed, statistical performance, generalization, and bias.
In this work, we propose a new class of EOT solvers (\algo),
that can estimate both plans and transport maps.
We take advantage of several opportunities to optimize the computation of EOT solutions by {\em dividing} mass displacement using a time discretization, borrowing inspiration from dynamic OT formulations~\citep{mccann1997convexity}, and {\em conquering} each of these steps using EOT with properly scheduled parameters. We provide experimental evidence demonstrating that \algo is a faster and more robust alternative to EOT solvers when computing couplings and maps at large scales, even outperforming neural network-based approaches. We also prove the statistical consistency of \algo when estimating OT maps. 
\looseness-1
\end{abstract}

\section{Introduction}\label{sec:intro}
Many problems in generative machine learning and natural sciences---notably biology~\citep{schiebinger2019optimal,bunne2023}, astronomy~\citep{metivier2016measuring} or quantum chemistry~\citep{PhysRevA.85.062502}---require aligning datasets or learning to map data points from a source to a target distribution. These problems stand at the core of optimal transport theory~\citep{santambrogio2015optimal} and have spurred the proposal of various solvers~\citep{peyre2019computational} to perform these tasks reliably. In these tasks, we are given $n$ and $m$ points respectively sampled from source and target probability distributions on $\mathbb{R}^d$, with the goal of either returning a coupling \emph{matrix} of size $n\times m$ (which solves the so-called \citeauthor{Kan42} problem), or a vector-valued \emph{map estimator} that extends to out-of-sample data (solving the \citeauthor{Monge1781} problem).

In modern applications, where $n,m\gtrsim 10^4$, a popular approach to estimating either coupling or maps is to rely on a regularization of the original~\citeauthor{Kan42} linear OT formulation using neg-entropy. This technique, referred to as \emph{entropic OT}, can be traced back to \citeauthor{schrodinger1931umkehrung} and was popularized for ML applications in \citep{cuturi2013sinkhorn} (see \cref{sec:background}). Crucially, EOT can be solved efficiently with \citeauthor{sinkhorn1964relationship}'s algorithm (\cref{algo:sinkhorn}), with favorable computational~\citep{AltWeeRig17,JMLR:v23:20-277} and statistical properties~\citep{genevay2019entropy,mena2019statistical} compared to linear programs. Most couplings computed nowadays on large point clouds within ML applications are obtained using EOT solvers that rely on variants of the~\citeauthor{sinkhorn1964relationship} algorithm, whether explicitly, or as a lower-level subroutine~\citep{pmlr-v139-scetbon21a,pmlr-v162-scetbon22b}. 
The widespread adoption of EOT has spurred many modifications of \citeauthor{sinkhorn1964relationship}'s original algorithm (e.g., through acceleration~\citep{thibault2021overrelaxed} or initialization~\citep{pmlr-v206-thornton23a}), and encouraged its incorporation within neural-network OT approaches~\citep{pooladian2023multisample,tong2023improving,uscidda2023monge}.

\begin{figure}[t]
    \centering
\includegraphics[width= \linewidth]{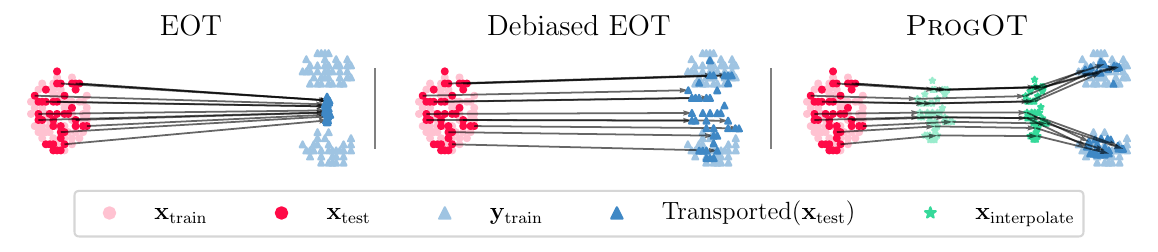}
    \caption{\textit{(left)} EOT solvers collapse when the value of $\varepsilon$ is not properly chosen. This typically results in biased map estimators and in blurry couplings (see Fig.~\ref{fig:couplings} for the coupling matrix obtained between $\vx_{\text{train}}$ and $\vy_{\text{train}}$). \textit{(middle)} Debiasing the output of EOT solvers
    can prevent a collapse to the mean seen in EOT estimators, but computes the same coupling. 
    \algo \textit{(right)} ameliorates these problems in various ways: 
    by decomposing the resolution of the OT problem into multiple time steps, and using various forms of progressive scheduling, we recover {\em both} a coupling whose entropy can be \emph{tuned} automatically and a map estimator that is fast and reliable.}
    \vspace{-10pt}
    \label{fig:figure1}
\end{figure}
\begin{wrapfigure}{r}{.45\textwidth}
\vspace{-18pt}
\centering \includegraphics[width=.45\textwidth]{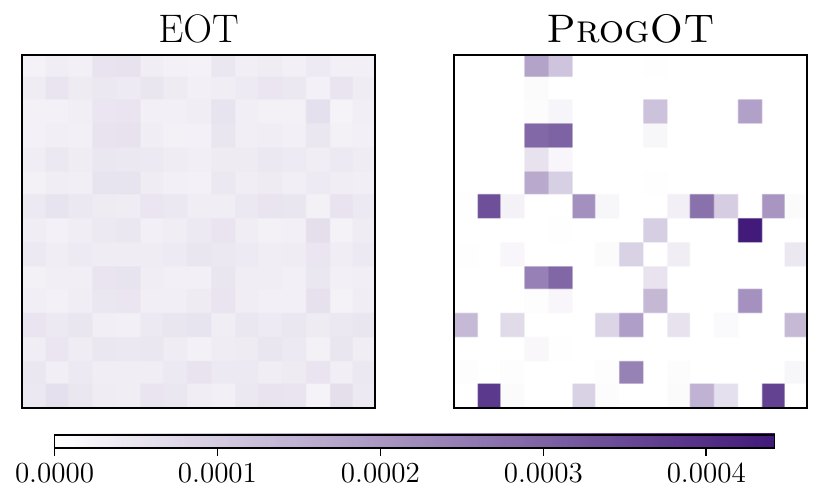}
    \caption{Coupling matrices between  train points in Fig.~\ref{fig:figure1}. Comparison of EOT with a fairly large $\varepsilon$, and \algo which automatically tunes the entropy of its coupling according to the target point cloud's dispersion.\looseness-1}
    \label{fig:couplings}
    \vspace{-2em}
\end{wrapfigure}
Though incredibly popular, \citeauthor{sinkhorn1964relationship}'s algorithm is not without its drawbacks.  While a popular tool due its scalability and simplicity, its numerical behavior is deeply impacted by the amount of neg-entropy regularization, driven by the hyperparameter $\varepsilon$. Some practitioners suggest to have the parameter nearly vanish~\citep{pmlr-v115-xie20b,schmitzer2019stabilized}, others consider the case where it diverges, highlighting links with the maximum mean discrepancy~\citep{ramdas2017wasserstein,genevay2019sample}. \looseness-1

Several years after its introduction to the machine learning community~\citep{cuturi2013sinkhorn}, choosing a suitable regularization term for EOT remains a thorny pain point. 
Common approaches are setting $\eps>0$ to a default value (e.g., the max~\citep{flamary2021pot} or mean~\citep{ott-jax} normalization of the transport cost matrix), 
incorporating a form of cross-validation or an unsupervised criterion~\citep{vacher2022parameter,van2023optimal}, or scheduling $\varepsilon$~\citep{mvr,feydy2020geometric}. When $\eps$ is too large, the algorithm converges quickly, but yields severely biased maps (\cref{fig:figure1}, left), or blurry, uninformative couplings  (\cref{fig:couplings}). Even theoretically and numerically \emph{debiasing} the \citeauthor{sinkhorn1964relationship} solver (\cref{fig:figure1}, middle) does not seem to fully resolve the issue \citep{feydy2019interpolating,pooladian2022debiaser}. To conclude, while strategies exist to alleviate this bias, there currently exists no one-size-fits-all solution to this problem. \looseness-1

\textbf{Our contribution: an EOT solver with a dynamic lens.} 
Recent years have witnessed an explosion in neural-network approaches based on the so-called ~\citeauthor{benamou2000computational} dynamic formulation of OT~\citep{lipman2022flow,liu2022rectified,tong2023improving,pooladian2023multisample}. 
A benefit of this perspective is the ability to split the OT problem into simpler sub-problems that are likely better conditioned than the initial transport problem. With this observation, we propose a novel family of \emph{progressive} EOT solvers, called \algo, that are meant to be sturdier and easier to parameterize than existing solvers. Our key idea is to exploit the dynamic nature of the problem, and vary parameters \textit{dynamically}, such as $\varepsilon$ and convergence thresholds, along the transport. We show that \algo
\vspace{-1mm}
\begin{itemize}[leftmargin=.3cm,itemsep=.05cm,topsep=0cm,parsep=2pt]
    \item can be used  to recover both \citeauthor{Kan42} couplings and~\citeauthor{Monge1781} map estimators,
    \item gives rise to a novel, provably statistically consistent map estimator under standard assumptions.
    \item strikes the right balance between computational and statistical tradeoffs,
    \item can outperform other (including neural-based) approaches on real datasets,
\end{itemize} 

\section{Background}\label{sec:background}

\textbf{Optimal transport.}
For domain $\Omega \subseteq \R^d$, let $\calP_2(\Omega)$ denote the space of probability measures over $\Omega$ with a finite second moment, and let $\calP_{2,\text{ac}}(\Omega)$ be those with densities. Let $\mu, \nu \in \calP_2(\Omega)$, and let $\Gamma(\mu,\nu)$ be the set of joint probability measures with left-marginal $\mu$ and right-marginal $\nu$. We consider a translation invariant cost function $c(x,y):=h(x-y)$, with $h$ a strictly convex function, and define the Wasserstein distance, parameterized by $h$, between $\mu$ and $\nu$
\begin{align}\label{eq:wassdist}
    W(\mu,\nu) \defeq \inf_{\pi \in \Gamma(\mu,\nu)}\iint h(x-y)\dd\pi(x,y)\,.
\end{align}
This formulation is due to \citet{Kan42}, and we call the minimizers to \eqref{eq:wassdist} \emph{OT couplings} or OT plans, and denote it as $\pi_0$.  
A subclass of couplings are those induced by \emph{pushforward maps}.  We say that $T:\R^d\to\R^d$ pushes  $\mu$ forward to $\nu$ if $T(X)\sim\nu$ for $X \sim \mu$, and write $T_\# \mu=\nu$. Given a choice of cost, we can define the \citet{Monge1781} formulation of OT\looseness-1
\begin{align}\label{eq:monge}
    T_0 \coloneqq \argmin_{T: T_{\#}\mu=\nu}\int h(x - T(x))\dd\mu(x)\,
\end{align}
where the minimizers are referred to as \emph{\citeauthor{Monge1781} maps}, or OT maps from $\mu$ to $\nu$. 
Unlike OT couplings, OT maps are not always guaranteed to exist.
Though, if $\mu$ has a density, we obtain significantly more structure on the OT map:
\begin{theorem}[\citeauthor{Bre91}'s Theorem \citeyearpar{Bre91}]\label{thm:brenier_thm} Suppose $\mu \in \calP_{2,\text{ac}}(\Omega)$ and $\nu \in \calP_2(\Omega)$. Then there exists a unique solution to \eqref{eq:monge} that is of the form $T_0 = \Id - \nabla h^* \circ \nabla f_0$, where $h^*$ is the convex-conjugate of $h$, i.e. $h^*(y)\coloneqq \max_x \langle x, y\rangle - h(x)$, and 
\begin{align}\label{eq:kant_dual}
    (f_0,g_0) \in \argmax_{(f,g)\in\calF} \int f \dd\mu + \int g \dd\nu\,,
\end{align}
where $\calF \coloneqq \{(f,g) \in L^1(\mu)\times L^1(\nu) : \ f(x)+g(y)\leq h(x-y),\, \forall x,y \in \Omega.\}$. Moreover, the OT plan is given by $\pi_0(\dd x, \dd y) =  \delta_{T_0(x)}(y)\mu(\dd x)$.
\end{theorem}
Importantly, \eqref{eq:kant_dual} is the \emph{dual} problem to \eqref{eq:wassdist} and the pair of functions $(f_0,g_0)$ are referred to as the \emph{optimal Kantorovich potentials}.
Lastly, we recall the notion of geodesics with respect to the Wasserstein distance. For a pair of measures $\mu$ and $\nu$ with OT map $T_0$, the \citeauthor{mccann1997convexity} interpolation between $\mu$ and $\nu$ is defined as
\begin{align}\label{eq:mccann}
    \mu_\alpha \defeq ((1-\alpha)\Id + \alpha T_0)_\#\mu\,,
\end{align}
where $\alpha\in[0,1]$. Equivalently, $\mu_\alpha$ is the law of 
$X_\alpha = (1-\alpha)X + \alpha T_0(X)$, where $X \sim \mu$.
In the case where $h = \|\cdot\|^p$ for $p > 1$, the \citeauthor{mccann1997convexity} interpolation is in fact a \emph{geodesic} in the Wasserstein space~\citep{ambrosio2005gradient}. While this equivalence may not hold for general costs, the \citeauthor{mccann1997convexity} interpolation still provides a natural path of measures between $\mu$ and $\nu$~\citep{liu2022rectified}.\looseness-1
\looseness-1

\vspace{5pt}
\begin{wrapfigure}{r}{0.5\textwidth}
\begin{minipage}{0.5\textwidth}
 \vspace{-25pt}
\begin{algorithm}[H]
\caption{\textsc{Sink}$(\ba, \bX, \bb, \bY, \varepsilon, \tau, \bff_{\text{init}}, \bg_{\text{init}})$.}
\label{algo:sinkhorn}
\begin{algorithmic}[1]
    \State{$\bff, \bg\leftarrow\bff_{\text{init}}, \bg_{\text{init}}$ (zero by default)}
    \State{$\bx_1,\dots, \bx_n=\bX, \quad \by_1, \dots, \by_m= \bY$}
    \State{$\mathbf{C}\leftarrow [h(\bx_i-\by_j)]_{ij}$}
    \While{$\|\exp\left(\tfrac{\mathbf{C}-\bff\oplus\bg}{\eps}\right)\mathbf{1}_m-\ba\|_1<\tau$}
        \State $\bff \leftarrow \varepsilon\log \ba - \min_\varepsilon(\mathbf{C}-\bff\oplus\bg) + \bff$
        \State $\bg \leftarrow \varepsilon\log \bb - \min_\varepsilon(\mathbf{C}^\top-\bg\oplus\bff) + \bg$
    \EndWhile
    \State{{\bfseries return} {$\bff, \bg, \mathbf{P}=\exp\left((\mathbf{C}-\bff\oplus\bg)/\eps\right)$ \label{lst:line:coupling}}}
\end{algorithmic}
\end{algorithm}
\end{minipage}
\vspace{-10pt}
\end{wrapfigure}

\paragraph{Entropic OT.}
Entropic regularization has become the de-facto approach to estimate all three variables $\pi_0$, $f_0$ and $T_0$ using samples $(\bx_1,\ldots,\bx_n)$ and $(\by_1,\ldots,\by_m)$, both weighted by probability weight vectors $\ba\in\mathbb{R}^n_+,\bb\in\mathbb{R}^m_+$ summing to $1$, to form approximations $\hat{\mu}_n = \sum_{i=1}^n \ba_i \delta_{\bx_i}$ and $\hat{\nu}_m = \sum_{i=1}^m \bb_j \delta_{\by_j}$. A common formulation of the EOT problem is the following $\eps$-strongly concave program:
\begin{align}\label{eq:entdual}
    \bff^\star,\bg^\star = \argmax_{\substack{ \bff \in \R^n, \bg \in \R^m }} & \langle \bff,\ba \rangle + \langle \bg, \bb\rangle\notag\\
    & - \eps \langle e^{\bff/\eps},\mathbf{K}e^{\bg/\eps}\rangle\,,
\end{align}
where $\varepsilon>0$ and $\mathbf{K}_{i,j} = [\exp(-(\bx_i - \by_j)/\eps)]_{i,j} \in \R^{n\times m}_+$. 
We can verify that \eqref{eq:entdual} is a regularized version of \eqref{eq:kant_dual} when applied to empirical measures \citep[Proposition 4.4]{peyre2019computational}. 
\citeauthor{sinkhorn1964relationship}'s algorithm presents an iterative scheme for obtaining $(\bff^\star,\bg^\star)$, and we recall it in Algorithm~\ref{algo:sinkhorn}, where for a matrix  
$\mathbf{S}=[\mathbf{S}_{i,j}]$ we use the notation $\min_\varepsilon(\mathbf{S}) \coloneqq [-\varepsilon \log\left( \mathbf{1}^\top e^{-\mathbf{S}_{i,\cdot}/\varepsilon}\right)]_i$, and $\oplus$ is the tensor sum of two vectors, i.e. $\bff\oplus \bg := [\bff_i+\bg_j]_{ij}.$
Note that solving \eqref{eq:entdual} also outputs a valid coupling $\mathbf{P}^\star_{i,j} = \mathbf{K}_{i,j} \exp(-(\bff^\star _i+\bg^\star_j)/\eps)$, which approximately solves the finite-sample counterpart of \eqref{eq:wassdist}.
Additionally, the optimal potential $f_0$ can be approximated by the {\em entropic potential}\looseness-1
\begin{align}\label{eq:entpot}
    \hat{f}_\eps(x) \defeq \mine([\bg^\star_j - h(x - \by_j)]_j),
\end{align}
where an analogous expression can be written for $\hat{g}_\eps$ in terms of $\bff^\star$. 
Using the entropic potential, we can also approximate the optimal transport map $T_0$
by the \emph{entropic map}
\begin{equation}\label{eq:Tent}
\hat{T}_{\eps}(x) = x - \nabla h^*\circ \nabla\hat{f}_\eps(x)\,.
\end{equation}
This connection is shown in \citet[Proposition 2]{pooladian2021entropic} for $h=\tfrac12\|\cdot\|^2$ and \citep{cuturi2023monge} for more general functions. 

\section{Progressive Estimation of Optimal Transport}\label{sec:method}
We consider the problem of estimating the OT solutions $\pi_0$ and $T_0$, given empirical measures $\hat \mu$ and $\hat \nu$ from $n$~i.i.d. samples.
Our goal is to design an algorithm which is numerically stable, computationally light, and yields a consistent estimator. The entropic map \eqref{eq:Tent} is 
an attractive option to estimate OT maps compared to other consistent estimators \citep[e.g., ][]{hutter2021minimax,manole2021plugin}. 
In contrast to these methods, the entropic map is tractable since it is the output of \citeauthor{sinkhorn1964relationship}'s algorithm. 
While \cite{pooladian2021entropic} show that the entropic map is a \emph{bona fide} estimator of the optimal transport map, it hides the caveat that the estimator is always \emph{biased}. 
For any pre-set $\eps > 0$, the estimator is never a valid pushforward map i.e., \smash{$(\hat{T}_\eps)_\# \mu\ne\nu$}, and this holds true as the number of samples tends to infinity. In practice, the presence of this bias implies that the performance of \smash{$\hat{T}_\eps$} is sensitive to the choice of $\eps$, e.g. as in \cref{fig:figure1}.
Instead of having Sinkhorn as the end-all solver, we propose to use it as a \emph{subroutine}. Our approach is to \emph{iteratively} move the source closer to the target, thereby creating a sequence of matching problems that are \emph{increasingly easier} to solve. As a consequence, the algorithm is less sensitive to the choice of $\eps$ for the earlier EOT problems, since it has time to correct itself at later steps.
To move the source closer to the target, we construct a \citeauthor{mccann1997convexity}-type interpolator which uses the entropic map \smash{$\Tent$} of the previous iterate, as outlined in the next section. \looseness-1

\subsection{Method}
As a warm-up, consider $T_0$ the optimal transport map from $\mu$ to $\nu$. We let $T^{(0)} \defeq T_0$ and define $S^{(0)} \coloneqq (1-\alpha_0)\Id + \alpha_0 T^{(0)}$. This gives rise to the measure \smash{$\mu^{(1)} = S^{(0)}_\#\mu$}, which traces out the \citeauthor{mccann1997convexity} interpolation between $(\mu, \nu)$ as $\alpha$ varies in the interval $(0, 1)$. Then, letting $T^{(1)}$ be the optimal transport map for the pair $(\mu^{(1)}, \nu)$, it is straightforward to show that $T^{(1)} \circ S^{(0)} = T^{(0)}$. In other words, in the idealized setting, composing the output of a progressive sequence of Monge problems along the \citeauthor{mccann1997convexity} interpolation path recovers the solution to the original Monge problem.

\begin{wrapfigure}{r}{.35\textwidth}
\vspace{-7pt}
    \includegraphics[width = 0.35\textwidth]{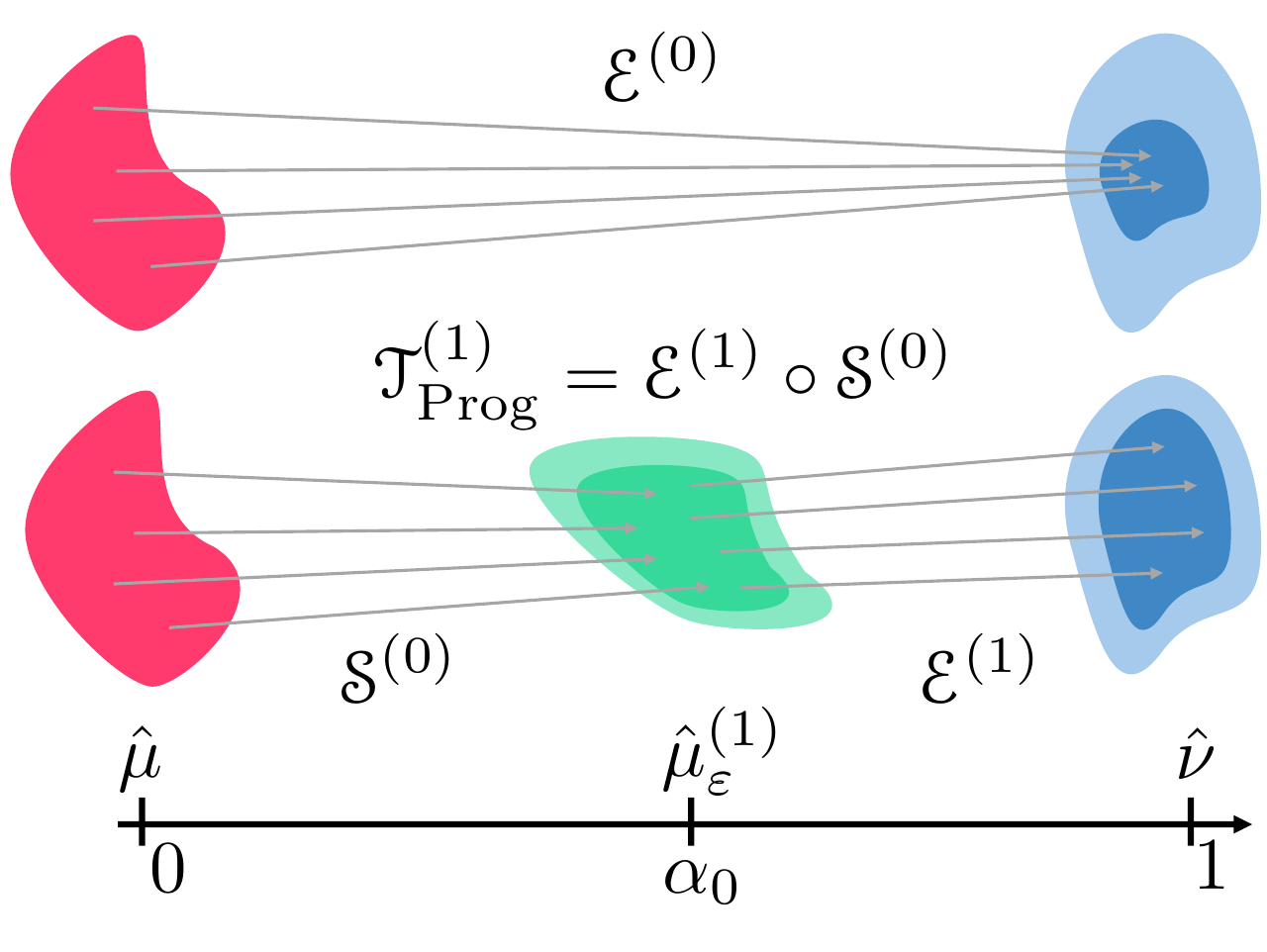}
    \caption{Intuition of \algo: By iteratively fitting to the interpolation path, the final transport step is less likely to collapse, resulting in more stable solver.}
    \label{fig:algo_visual}
    \vspace{-5pt}
\end{wrapfigure}
Building on this observation, we set up a progressive sequence of {\em entropic} optimal transport problems, along an {\em estimated} interpolation path, between the {\em empirical} counterparts of $(\mu, \nu)$. We show that, as long as we remain close to the true interpolation path (by not allowing $\alpha$ to be too large), the final output is close to $\nu$. Moreover, as the algorithm progresses, 
choosing the parameters $\eps_i$ becomes a less arduous task, and computation of \smash{$\Tent$} becomes a more stable numerical problem.

At step zero, we set \smash{$\muent\ith{0} = \hat\mu$} and calculate the entropic map \smash{$\myTent\ith{0}\defeq\hat{T}_{\eps_0}$} from samples $(\muent\ith{0},\hat \nu)$ with a regularization parameter $\eps_0>0$.
To set up the next EOT problem, we create an intermediate distribution via the \citeauthor{mccann1997convexity}-type interpolation
\[
\muent\ith{1} \defeq \myS\ith{0}_{\#} \muent\ith{0},\; \myS\ith{0}\defeq (1-\alpha_0)\Id + \alpha_0 \myTent\ith{0}\,,
\]
with $\alpha_0 \in (0, 1)$. In doing so, we are mimicking a step along the interpolation path for the pair $(\mu, \nu)$. In fact, we can show that \smash{$\muent\ith{1}$} is close to $\mu_{\alpha_0}$ as defined in \eqref{eq:mccann} (see \cref{lem:recursion2}). For the next iteration of the algorithm, we choose $\eps_1$ and $\alpha_1$, compute  
\smash{$\myTent^{(1)}$} the entropic map for the pair \smash{$(\muent\ith{1},\hat\nu)$} with regularization $\eps_1$, and move along the estimated interpolation path by computing the distribution \smash{$\muent\ith{2}$}. We repeat the same process for $K$ steps. The algorithm then outputs the {\em progressive entropic} map
\begin{equation*}%
   \Tprog\ith{K} \coloneqq \myTent\ith{K} \circ \myS\ith{K-1} \circ \dots \circ \myS\ith{0} \,,
\end{equation*}
where $\myS\ith{k} = (1-\alpha_{k})\Id + \alpha_{k} \myTent\ith{k}$ is the \citeauthor{mccann1997convexity}-type interpolator at step $k$.
\cref{fig:algo_visual} visualizes the one-step algorithm, and \cref{def:progot} formalizes the construction of our progressive estimators. \looseness-1
 
\begin{mydef}[\algo]\label{def:progot}
    For two empirical measures $\hat \mu,\hat \nu$, and given step and regularization schedules $(\alpha_k)_{k=0}^K$ and $(\eps_k)_{k=0}^K$, the \algo map estimator $\Tprog^{(K)}$ is defined as the composition 
    $$\Tprog^{(K)} \defeq \myTent^{(K)} \circ \myS\ith{K-1} \circ \dots \circ \myS\ith{0}$$
    where these maps are defined recursively, starting from $\hat{\mu}_\eps\ith{0} \coloneqq \hat{\mu}$, and then at each iteration: 
    \begin{itemize}[leftmargin=.3cm,itemsep=.0cm,topsep=0cm,parsep=2pt]
        \item $\myTent\ith{k}$ is the entropic map $\hat{T}_{\eps_k}$, computed between samples $(\hat{\mu}_\eps^{(k)}, \hat{\nu})$ with regularization $\eps_k$.
        \item $\myS\ith{k} := (1-\alpha_{k})\Id + \alpha_{k} \myTent^{(k)}$, is a \citeauthor{mccann1997convexity}-type interpolating map at time $\alpha_k$.
        \item $\hat{\mu}_\eps\ith{k+1} := \myS\ith{k}_{\#} \hat{\mu}_\eps\ith{k}$ the updated measure used in the next iteration.
    \end{itemize}
    Additionally, the \algo coupling matrix $\mathbf{P}$ between $\hat\mu$ and $\hat\nu$ is identified with the matrix solving the discrete EOT problem between $\hat{\mu}\ith{K}_\eps$ and $\hat{\nu}$.
\end{mydef}

The sequence of $(\alpha_k)_{k=0}^K$ characterizes the speed of movement along the path. By choosing $\alpha_k= \alpha(k)$ we can recover a constant-speed curve, or an accelerating curve which initially takes large steps and as it gets closer to the target, the steps become finer, or a decelerating curve which does the opposite. This is discussed in more detail in \cref{sec:implementation} and visualized in Figure~(\ref{fig:exploratory}-C).
Though our theoretical guarantee requires a particular choice for the sequence $(\eps_k)_{k=0}^K$ and $(\alpha_k)_{k=0}^K$, our experimental results reveal that the performance of our estimators is not sensitive to this choice.
We hypothesize that this behavior is due to the fact that \algo is ``self-correcting''---by steering close to the interpolation path, later steps in the trajectory can correct the biases introduced in earlier steps.

\subsection{Theoretical Guarantees}\label{sec:theory_guarantees}
By running \algo, we are solving a sequence of EOT problems, each building on the outcome of the previous one. Since error can potentially accumulate across iterations, it leads us to ask if the algorithm diverges from the interpolation path and whether the ultimate  progressive map estimator \smash{$T_{\mathrm{Prog}}\ith{K}$} is consistent,
focusing on the squared-Euclidean cost of transport, i.e.,  $h = \tfrac12\|\cdot\|^2$. . 
To answer this question, we assume\looseness-1
\begin{description}
    \item \textbf{(A1)} $\mu,\nu \in \calP_{2,\text{ac}}(\Omega)$ with $\Omega \subseteq \R^d$ convex and compact, with $0 < \nu_{\min} \leq \nu(\cdot) \leq \nu_{\max}$ and $\mu(\cdot) \leq \mu_{\max}$, 
    \item \textbf{(A2)} the inverse mapping $x \mapsto (T_0(x))^{-1}$ has at least three continuous derivatives, 
    \item \textbf{(A3)} there exists $\lambda,\Lambda > 0$ such that $\lambda I \preceq D T_0(x) \preceq \Lambda I$, for all $x \in \Omega$ ($D$ denotes Jacobian)
\end{description}
and prove that \algo yields a consistent map estimator.
Our error bound depends on the number of iterations $K$, via a constant multiplicative factor. 
Implying that \smash{$T_{\mathrm{Prog}}\ith{K}$} is consistent as long as $K$ does not grow too quickly as a function of $n$ the number of samples. In experiments, we set $K\ll n$. \looseness-1

\begin{theorem}[Consistency of Progressive Entropic Maps]\label{thm:consistency_multistep}
    Let $h=\tfrac12\|\cdot\|^2$.
    Suppose $\mu,\nu$ and their optimal transport map $T_0$ satisfy \textbf{(A1)}-\textbf{(A3)}, and further suppose we have $n$ i.i.d.~samples from both $\mu$ and $\nu$. Let $\Tprog\ith{k}$ be as defined in \cref{def:progot}, with parameters $\eps_k \asymp n^{-\frac{1}{2d}}, \alpha_k \asymp n^{\frac{-1}{d}}$ for all $k \in [K]$. Then, the progressive entropic map is consistent and converges to the optimal transport map as\looseness-1
    \begin{align*}
    \Exp \norm{\Tprog\ith{K} - T_0}^2_{L^2(\mu)} \lesssim_{\log(n), K}  n^{-\frac{1}{d}}\,
 \end{align*}
 where the notation implies that the inequality ignores terms of rate $\mathrm{Poly}(\log(n), K)$.
\end{theorem}

The rate of convergence for \algo is slower than the convergence of entropic maps shown by \citet{pooladian2021entropic} under the same assumptions, with the exception of convexity of $\Omega$. However, the rates that \cref{thm:consistency_multistep} suggests for the parameters $\alpha_k$ and $\eps_k$ are set merely to demonstrate convergence and do not reflect how each parameter should be chosen as a function of $k$ when executing the algorithm. We will present practical schedules for $(\alpha_k)_{k=1}^K$ and $(\eps_k)_{k=1}^K$ in \cref{sec:implementation}. The proof is deferred to \cref{app:proofs}; here we present a brief sketch.\looseness-1
\begin{proof}[Proof sketch]
In \cref{lem:summation_lemma}, we show that
\begin{align*}
    \Exp \norm{\Tprog\ith{K} - T_0}^2_{L^2(\mu)} \lesssim \sum_{k=0}^K \Delta_k \defeq \sum_{k=0}^K \E \|\myTent^{(k)} - T^{(k)}\|^2_{L^2(\mu^{(k)})}\,,
\end{align*}
where $\mu^{(k)}$ is a point on the true interpolation path, and $T^{(k)}$ is the optimal transport map emanating from it. Here, $\myTent^{(k)}$ is the entropic map estimator between the final target points and the data that has been pushed forward by earlier entropic maps. It suffices to control the term $\Delta_k$. 
Since $\myTent^{(k)}$ and $T^{(k)}$ are calculated for {\em different} measures, 
we prove a novel stability property (\cref{prop:stability_phi}) to show that
 along the interpolation path, these intermediate maps remain close to their unregularized population counterparts, if $\alpha_k$ and $\eps_k$ are chosen as prescribed.
 This result is based off the recent work by \citet{divol2024tight} and allows us to recursively relate the estimation at the $k$-th iterate to the estimation at the previous ones, down to $\Delta_0$. Thus, \cref{lem:recursion2} tells us that, under our assumptions and parameter choices $\alpha_k \asymp n^{-1/d}$ and $\eps_k \asymp n^{-1/2d}$, it holds that for all $k \geq 0$\looseness-1
\begin{align*}
\Delta_k \lesssim_{\log(n)} n^{-1/d}\,.
\end{align*}
Since the stability bound allows us to relate $\Delta_k$ to $\Delta_0$, combined with the above, we have that
\begin{align*}
    \Delta_k \lesssim_{\log(n)} \Delta_0 \lesssim_{ \log(n)} n^{-1/d}\,, 
\end{align*}
where the penultimate inequality uses the existing estimation rates from \cite{pooladian2021entropic}, with our parameter choice for $\eps_0$.
\end{proof}

\begin{proposition}[Stability of entropic maps with variations in the source measure]\label{prop:stability_phi}
Let $h=\tfrac12\|\cdot\|^2$.
Let $\mu, \mu', \rho$ be probability measures over a compact domain with radius $R$. Suppose $T_\eps, T_\eps'$ are, respectively, the entropic maps from $\mu$ to $\rho$ and $\mu'$ to $\rho$, both with the parameter $\eps > 0$. 
    Then \begin{align*}
        \norm{T_\eps - T_\eps'}^2_{L^2(\mu)} \leq 3R^2 \eps^{-1} \WD_2^2(\mu, \mu')\,.
    \end{align*}
\end{proposition}

\section{Computing Couplings and Map Estimators with \algo}\label{sec:implementation}

Following the presentation and motivation of \algo in \cref{sec:method}, here we outline a practical implementation.
Recall that $\hat{\mu}_n= \sum_{i=1}^n \ba_i \delta_{\bx_i}$ and $\hat{\nu}_m = \sum_{j=1}^m \bb_j \delta_{\by_j}$, and we summarize the locations of these measures to the matrices $\bX=(\bx_1,\dots,\bx_n)$ and \smash{$\bY=(\by_1,\dots,\by_m)$}, which are of size $n\times d$ and $m\times d$, respectively.
Our \algo solver, concretely summarized in \Cref{alg:progot}, takes as input two weighted point clouds, step-lengths $(\alpha_k)_k$, regularization parameters $(\eps_k)_k$, and threshold parameters $(\tau_k)_k$, to output two objects of interest: the final coupling matrix $\mathbf{P}$ of size $n\times m$, as illustrated in Figure~\ref{fig:couplings}, and the entities needed to instantiate the $\Tprog$ map estimator, where an implementation is detailed in Algorithm~\ref{alg:progotmap}. We highlight that \cref{alg:progot} incorporates a warm-starting method when instantiating  \citeauthor{sinkhorn1964relationship} solvers (Line 3). This step may be added to improve the runtime.\looseness-1

\begin{tabular}{@{}c@{}c@{}}
\begin{minipage}[t]{0.56\textwidth}
\vspace{-12pt}
\begin{algorithm}[H]
\caption{\algo\label{alg:progot}\smash{$\!\!(\ba, \bX, \bb, \bY, (\eps_k, \alpha_k, \tau_k)_k)$}}
\begin{algorithmic}[1]
\State $\bff=\mathbf{0}_n, \bg^{(-1)}=\mathbf{0}_m.$
\For{$k = 0, \dots, K$}
\State $\bff_{\text{init}}, \bg_{\text{init}}\leftarrow (1-\alpha_k)\,\bff, (1-\alpha_k)\,\bg^{(k-1)}$
\State $\bff, \bg^{(k)}, \mathbf{P} \leftarrow \text{Sink}(\ba, \bX, \bb, \bY, \eps_k, \tau_k, \bff_{\text{init}}, \bg_{\text{init}})$
\State $\mathbf{Q}\leftarrow \diag(1/\mathbf{P}\mathbf{1}_m)\mathbf{P}$\label{lst:renorm}
\State $\mathbf{Z} \leftarrow [\nabla h^*(\sum_j \mathbf{Q}_{ij} \nabla h(\bx_i-\by_j))]_i \in\mathbb{R}^{n\times d}$
\State $\bX \leftarrow \bX - \alpha_{k}\,\mathbf{Z}$
\EndFor
\State \textbf{return:} Coupling matrix $\mathbf{P}$,
\State Map estimator \smash{$\Tprog[\bb,\bY,(\bg^{(k)},\eps_k, \alpha_k)_k](\cdot)$ }
\end{algorithmic}
\end{algorithm}
\end{minipage}
&
\,\,\,\begin{minipage}[t]{0.42\textwidth}
\vspace{-12pt}
\begin{algorithm}[H]
\caption{\smash{$\Tprog[\bb,\bY,(\bg^{(k)},\eps_k, \alpha_k)_k]$}\label{alg:progotmap}}
\begin{algorithmic}[1]
\State \textbf{input:} Source point $\bx\in\mathbb{R}^d$
\State \textbf{initialize:} $\by=\bx$, $\alpha_K$ reset to $1$.
\For{$k = 0, \dots K$}
\State $\mathbf{p} \leftarrow [\bb_j\exp(\tfrac{\bg^{(k)}-h(\by-\by_j)}{\eps_k})]_j$
\State \smash{$\mathbf{p} \leftarrow \mathbf{p}/ \mathbf{1}_m^T\mathbf{p}\in\mathbb{R}^m$}
\State $\mathbf{\Delta} \leftarrow [\nabla h(\by-\by_j)]_j\in\mathbb{R}^{m\times d}$
\State \smash{$\mathbf{z} \leftarrow \nabla h^*(\mathbf{p}^T\mathbf{\Delta})\in\mathbb{R}^d$}
\State $\by \leftarrow \by - \alpha_{k} \mathbf{z}.$
\EndFor
\State \textbf{return:} $\by$ 
\end{algorithmic}
\end{algorithm}
\end{minipage}
\end{tabular}

\textbf{Setting step lengths.} We propose three scheduling schemes for $(\alpha_k)_k$: decelerated, constant-speed and accelerated.
Let $t_k \in [0, 1]$ denote the progress along the interpolation path at iterate $k$. At step zero, $t_0 = \alpha_0$. Then at the next step, we progress by a fraction $\alpha_1$ of the remainder, and therefore $t_1 = t_0 + \alpha_1(1-\alpha_0)$. It is straightforward to show that
\smash{$
    t_k = 1-\prod_{\ell=1}^k(1-\alpha_\ell).
$}
We call a schedule constant speed, if $t_{k+1} - t_{k}$ is a constant function of $k$, whereas an accelerated (resp. decelerated) schedule has $t_{k+1} - t_{k}$ increasing (resp. decreasing) with $k$. \cref{tab:alpha_sched} presents the specific choices of $\alpha_k$ for each of these schedules. By convention, we set the last step to be a complete step, i.e., $\alpha_K=1$.\looseness-1

\textbf{Setting regularization schedule.} To set the regularization parameters $(\eps_k)_{k=0}^K$, we propose \cref{alg:eps_sched}. 
To set $\eps_0$, we use the average of the values in the cost matrix $[h(\bx_i-\by_j)]_{ij}$ between source and target, multiplied by a small factor, as implemented in \citep{ott-jax}.
Then for $\eps_K$, we make the following key observation. As the last iteration, the algorithm is computing $\myTent\ith{k}$, an entropic map roughly between the target measure and \emph{itself}. For this problem, we know trivially that the OT map should be the identity.
Therefore, given a set of values to choose from, we pick $\eps_K$ to be that which minimizes this error over a hold-out evaluation set of $\mathbf{Y}_{\mathrm{test}}=(\tilde \vy_j)_{j=1}^m$
\begin{wrapfigure}{r}{0.45\textwidth}
\begin{minipage}{0.45\textwidth}
\vskip-.3cm
\begin{algorithm}[H]
\caption{$\eps$-scheduler$(\mY_{\text{test}}, (s_p)_p,\beta_0)$ \label{alg:eps_sched}}
\begin{algorithmic}[1]
\State \textbf{recover} $\ba, \mX, \bb,\mY, (\alpha_k)_{k}$.
\State {\textbf{set} $\eps_0 \leftarrow \tfrac1{20}\frac{1}{nm} \sum_{ij} h(\mathbf{x}_i-\mathbf{y}_j)$.\label{lst:eps0}}
\State \textbf{set} $\sigma \leftarrow \tfrac1{20}\frac{1}{m^2} \sum_{lj} h(\mathbf{y}_l-\mathbf{y}_j)$.
\For{$p = 1, \dots, 
P$}
\State $\eps\leftarrow s_p \times\sigma.$
\State $\_, \bg, \_ \leftarrow \textsc{Sink}(\bb, \bY, \bb, \bY, \eps, \tau)$
\State $T = \Tprog[\bb, \bY, (\bg, \eps, 1)]$
\State $\error_p \leftarrow \|\mY_{\text{test}} - T(\mY_{\text{test}})\|^2$%
\EndFor
\State $p^\star=\argmin_{p} \mathrm{error}_p$.
\State $\eps_1 \leftarrow s_{p^\star} \times \sigma$. 
\State $(t_k)_k = (1 - \prod_{\ell=1}^k(1-\alpha_\ell))_k$.
\State \textbf{return:} $((1-t_k)\beta_0 \eps_0 + t_k \eps_1)_k$. 
\end{algorithmic}
\end{algorithm}
\end{minipage}
 \vskip-4mm
\end{wrapfigure}
\begin{equation*}
    \error(\eps; Y_{\mathrm{test}}) \coloneqq \sum_{j=1}^m\, \norm{\tilde \vy_j - \myTent\ith{K}(\tilde\vy_j)}_2^2. 
\end{equation*}
The intermediate values are then set by interpolating between $\beta_0\eps_0$ and $\eps_K$, according to the times $t_k$. \cref{fig:exploratory}-C visualizes the effect of applying \cref{alg:eps_sched} for scheduling, as opposed to choosing default values for $\eps_k$. \looseness-1

\textbf{Setting threshold schedule.} By setting the \citeauthor{sinkhorn1964relationship} stopping threshold $\tau_k$ as a function of time $k$, one can modulate the amount of compute effort spent by the \citeauthor{sinkhorn1964relationship} subroutine at each step. This can be achieved by decreasing $\tau_k$ linearly w.r.t. the iteration number, from a loose initial value, e.g., $0.1$, to a final target value $\tau_K \ll 1$. Doing so results naturally in sub-optimal couplings $\mathbf{P}$ and dual variables $g^{(k)}$ at each step $k$, which might hurt performance. However, two comments are in order: 
\textit{(i)} Because the \textit{last} threshold $\tau_K$ can be set independently, the final coupling matrix $\mathbf{P}$ returned by \algo can be arbitrarily feasible, in the sense that its marginals can be made arbitrarily close to $\ba,\bb$ by setting $\tau_K$ to a small value. 
This makes it possible to compare in a fair way to a direct application of the \citeauthor{sinkhorn1964relationship} algorithm.
\textit{(ii)} Because the coupling is normalized by its own marginal in line~\eqref{lst:renorm} of Algorithm~\ref{alg:progot}, we ensure that the barycentric projection computed at each step remains valid, i.e., the matrix $\mathbf{Q}$ is a transition kernel, with line vectors in the probability simplex. \looseness-1

\section{Experiments}\label{sec:experiments}
We run experiments to evaluate the performance of \algo across various datasets, on its ability to act as a map estimator, and to produce couplings between the source and target points. The code for \algo, is included in the OTT-JAX package \citep{ott-jax}.

\subsection{\algo as a Map Estimator} \label{sec:map_experiment}
\begin{figure}[t]
    \centering
    \includegraphics[width = \linewidth]{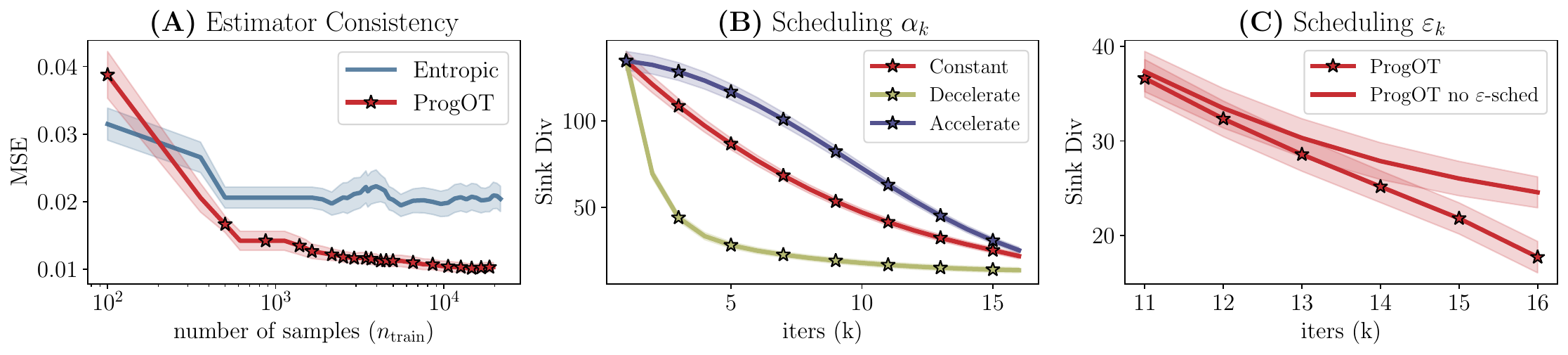}
    \vspace{-15pt}
    \caption{
    \textbf{(A)} Convergence of $\Tprog$ to the ground-truth map w.r.t. the empirical L2 norm, for $d = 4$. 
    \textbf{(B)} Effect of scheduling $\alpha_k$, for $d = 64$.
    \textbf{(C)} Effect of scheduling $\eps_k$ using \Cref{alg:eps_sched}, for $d = 64$.\looseness-1
    \label{fig:exploratory}}
    \vspace{-10pt}
\end{figure}

In map experiments, unless mentioned otherwise, we run \algo for $K=16$ steps, with a constant-speed schedule for $\alpha_k$, and the regularization schedule set via \cref{alg:eps_sched} with $\beta_0 = 5$ and $s_p \in \{2^{-3}, \dots, 2^3\}$.
In these experiments, we fit the estimators on training data using the $\ell^2_2$ transport cost,
and report their performance on test data in \cref{fig:exploratory} and \cref{tab:cell_benchmarks}.
To this end, we quantify the distance between two test point clouds ($\mX, \mY$) with the \emph{Sinkhorn divergence} \citep{2017-Genevay-AutoDiff,feydy2019interpolating}, always using the $\ell^2_2$ transport cost. Writing $\mathrm{OT}_\eps(\mX, \mY)$ for the objective value of \cref{eq:entdual}, the Sinkhorn divergence reads
\begin{equation}\label{eq:sink_div}
   D_{\eps_D}(\mX, \mY) \coloneqq \mathrm{OT}_{\eps_D}(\mX, \mY)-\tfrac{1}{2}\bigl(\mathrm{OT}_{\eps_D}(\mX, \mX)+\mathrm{OT}_{\eps_D}(\mY, \mY)\bigr) \,,
\end{equation}
where $\eps_D$ is 5\% of the mean (intra) cost seen within the target distribution (see \cref{app:experiment_details}).

\textbf{Exploratory Experiments on Synthetic Data.} 
We consider a synthetic dataset where $\mX$ is a $d$-dimensional point cloud sampled from a 3-component Gaussian mixture. The ground-truth $T_0$ is the gradient of an input convex neural network (ICNN) previously fitted to push roughly $\mX$ to a mixture of $10$ Gaussians \citep{korotin2021neural}. From this map, we generate the target point cloud $\mY$. 
Unless stated otherwise, we use $n_\mathrm{train} = 7000$ samples to train a progressive map between the source and target point clouds and visualize some of its properties in \cref{fig:exploratory} using $n_{\mathrm{test}} = 500$ test points.

\cref{fig:exploratory}-\textbf{(A)} demonstrates the convergence of $\Tprog$ to the true map as the number of training points grows, in empirical L2 norm, that is,
\smash{$
 \mathrm{MSE} = \tfrac{1}{n_{\mathrm{test}}}\sum_{i=1}^{n_{\mathrm{test}}}\norm{T_0(\vx_i) - \Tprog(\vx_i)}_2^2
$}.
\cref{fig:exploratory}-\textbf{(B)} shows the progression of \algo from source to target as measured by $D_{\eps_D}(\mX\ith{k}, \mY)$ where $\mX\ith{k}$ are the intermediate point clouds corresponding to \smash{$\hat \mu_\eps\ith{k}$}.
The curves reflect the speed of movement for three different schedules, e.g., the decelerated algorithm takes larger steps in the first iterations, resulting in $D_{\eps_D}(\mX\ith{k}, \mY)$ to initially drop rapidly.
Across multiple evaluations, we observe that the $\alpha_k$ schedule has little impact on performance and settle on the constant-speed schedule.
Lastly, \cref{fig:exploratory}-\textbf{(C)} plots $D_{\eps_D}(\mX\ith{k}, \mY)$ for the last $6$ steps of the progressive algorithm under two scenarios.
\algo uses regularization parameters set according to \cref{alg:eps_sched}, and \algo without scheduling, sets every $\eps_k$ as $5\%$ of the mean of the cost matrix between the point clouds of \smash{$(\mX\ith{k}, \mY)$}. This experiment shows that \cref{alg:eps_sched} can result in displacements \smash{$\mX\ith{k}$} that are ``closer'' to the target $\mY$, potentially improving the overall performance.

\textbf{Comparing Map Estimators on Single-Cell Data.} 
We consider the sci-Plex single-cell RNA sequencing data from \citep{srivatsan2020massively} which contains the responses of cancer cell lines to 188 drug perturbations, as reflected in their gene expression.
Visualized in \cref{fig:scrna_distrib}, we focus on $5$ drugs (Belinostat, Dacinostat, Givinostat, Hesperadin, and Quisinostat) which have a significant impact on the cell population as reported by \citet{srivatsan2020massively}. 
We remove genes which appear in less than 20 cells, and discard cells which have an incomplete gene expression of less than 20 genes, obtaining $n\approx 10^4$ source and $m\approx 500$ target cells, depending on the drug.
We whiten the data, take it to $\log(1+x)$ scale and apply PCA to reduce the dimensionality to $d=\{16, 64, 256\}$.
This procedure repeats the pre-processing steps of \citet{cuturi2023monge}.\looseness-1

We consider four baselines: (1) training an input convex neural network (ICNN) \citep{amos2017icnn} using the objective of \citet{amos2022amortizing} (2) training a feed-forward neural network regularized with the Monge Gap \citep{uscidda2023monge}, (3) instantiating the entropic map estimator \citep{pooladian2021entropic} and (4) its debiased variant \citep{feydy2019interpolating,pooladian2022debiaser}. 
The first two algorithms use neural networks, and we follow hyper-parameter tuning in~\citep{uscidda2023monge}.
We choose the number of hidden layers for both as $[128, 64, 64]$. For the ICNN we use a learning rate $\eta = 10^{-3}$, batch size $b= 256$ and train it using the Adam optimizer \citep{kingma2014adam} for $2000$ iterations.
For the Monge Gap we set the regularization constant $\lambda_{\mathrm{MG}} = 10$, $\lambda_{\mathrm{cons}} = 0.1$ and the \citeauthor{sinkhorn1964relationship} regularization to $\eps = 0.01$. We train the Monge Gap in a similar setting, except that we set $\eta = 0.01$.
To choose $\eps$ for entropic estimators, we split the training data to get an evaluation set and perform $5$-fold cross-validation on the grid of $\{2^{-3}, \dots, 2^3\}\times\eps_0$, where $\eps_0$ is computed as in line \ref{lst:eps0} of Algorithm~\ref{alg:eps_sched}.\looseness-1

We compare the algorithms by their ability to align the population of control cells, to cells treated with a drug.
We randomly split the data into $80\%-20\%$ train and test sets, and report the mean and standard error of performance over the {\em test set}, for an average of 5 runs. 
Detailed in \cref{tab:cell_benchmarks}, \algo outperforms the baselines consistently with respect to $D_{\eps_D}((\Tprog)_{\#}\mX, \mY)$. The table shows complete results for $3$ drugs, and the overall ranking based on performance across all $5$ drugs.
\cref{tab:gmm_benchmark} presents the synthetic counterpart to \cref{tab:cell_benchmarks}, using Gaussian Mixture data for $d=128, 256$.

\setlength{\tabcolsep}{0.2em}
\begin{table}[t]
    \centering
    \caption{Performance of \algo compared to baselines, w.r.t $\mathrm{D}_{\eps_D}$ between source and target of the sci-Plex dataset. Reported numbers are the average of 5 runs, together with the standard error.\looseness-1}
    \label{tab:cell_benchmarks}
{\scalebox{.84}{\begin{tabular}{|c|*{9}{c|}c|}
\hline
Drug & \multicolumn{3}{c|}{Belinostat} & \multicolumn{3}{c|}{Givinostat} & \multicolumn{3}{c|}{Hesperadin} & \multirow{2}{*}{\makecell{ 5-drug\\rank}}\\
\cline{1-10}
$d_\mathrm{PCA}$& 16 & 64 & 256 & 16 & 64 & 256 & 16 & 64 & 256& \\
\hline
\algo &\makecell{2.9$\pm$0.1} & 
\makecell{\textbf{8.8}$\pm$0.1} & 
\makecell{\textbf{20.8}$\pm$0.2} & 
\makecell{3.3$\pm$0.2} & 
\makecell{\textbf{9.0}$\pm$0.3} & 
\makecell{\textbf{21.9}$\pm$0.3} & 
\makecell{\textbf{3.7}$\pm$0.4} & 
\makecell{\textbf{10.1}$\pm$0.4} & 
\makecell{\textbf{23.1}$\pm$0.4} &  \textbf{1}\\
\hline
EOT & \makecell{\textbf{2.5}$\pm$0.1} & 
\makecell{9.6$\pm$0.1} & 
\makecell{22.8$\pm$0.2} & \makecell{3.9$\pm$0.4} & 
\makecell{10.0$\pm$0.1} & 
\makecell{24.7$\pm$0.9} &   \makecell{4.1$\pm$0.4} & 
\makecell{10.4$\pm$0.5} & 
\makecell{26$\pm$1.3} &  2 \\
\hline
\makecell{Debiased EOT} & \makecell{3.2$\pm$0.1} & 
\makecell{14.3$\pm$0.1} & 
\makecell{39.8$\pm$0.4} & \makecell{3.7$\pm$0.2} & 
\makecell{14.7$\pm$0.1} & 
\makecell{42.4$\pm$0.8} & \makecell{4.0$\pm$0.5} & 
\makecell{15.2$\pm$0.6} & 
\makecell{41$\pm$1.1} &  4\\
\hline
\makecell{Monge Gap} & \makecell{3.1$\pm$0.1} & 
\makecell{10.3$\pm$0.1} & 
\makecell{34.4$\pm$0.3} & 
\makecell{\textbf{2.8}$\pm$0.2} & 
\makecell{9.9$\pm$0.2} & 
\makecell{34.9$\pm$0.3} & \makecell{3.7$\pm$0.5}&\makecell{11.0$\pm$0.5} & \makecell{36$\pm$1.1}& 3 \\
\hline
ICNN & \makecell{5.0$\pm$0.1} & \makecell{14.7$\pm$ 0.1} & \makecell{42$\pm$1} & \makecell{5.1$\pm$0.1} & \makecell{14.8$\pm$0.2}& \makecell{40.3$\pm$0.1}& \makecell{4.0$\pm$0.4}& \makecell{14.4$\pm$0.5} & \makecell{46$\pm$2.1} & 5 \\
\hline
\end{tabular}}}
\vspace{-10pt}
\end{table}

\begin{figure}
    \centering
\includegraphics[width = \linewidth]{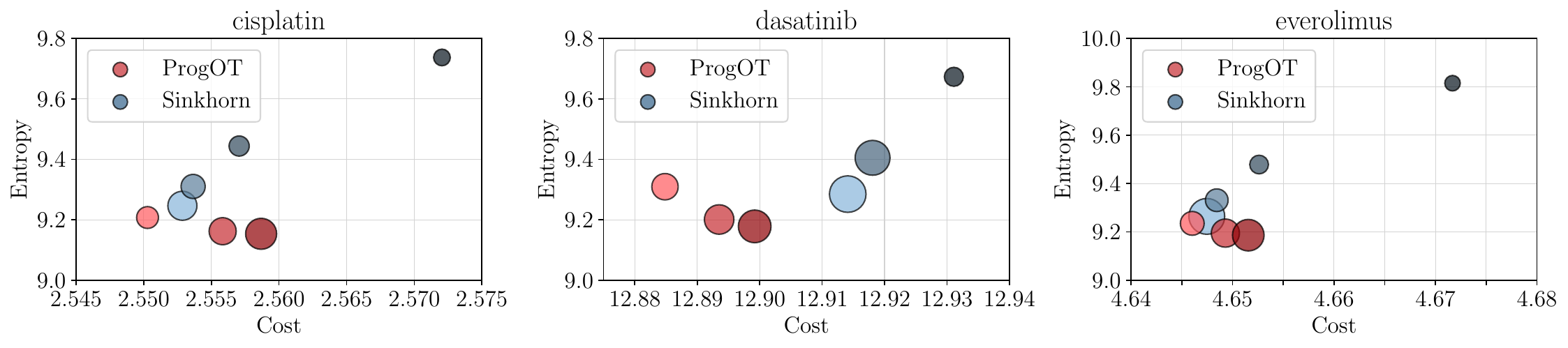}
\includegraphics[width = \linewidth]{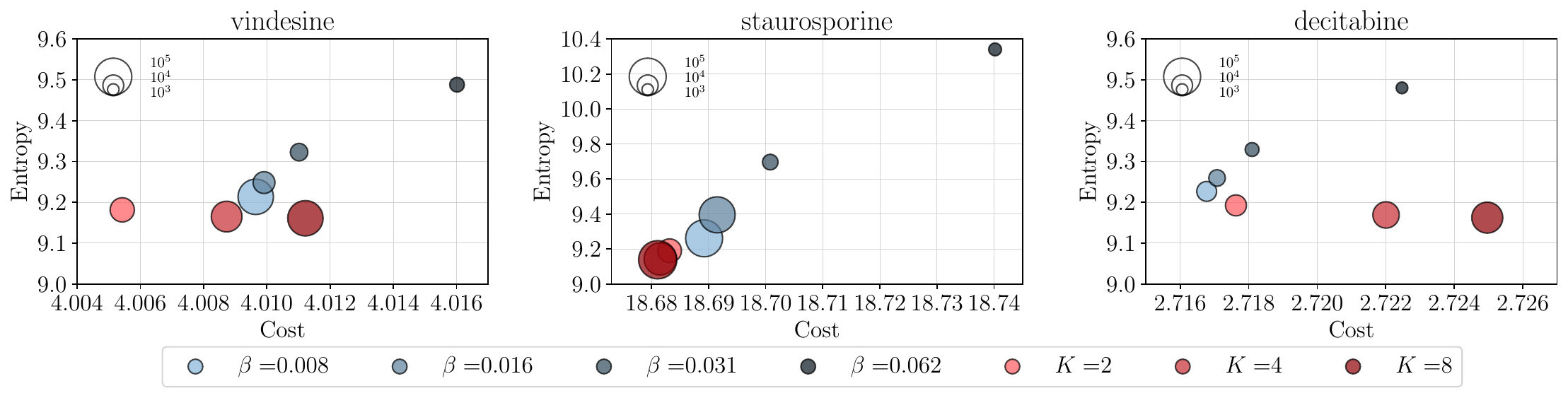}
    \caption{Performance as a coupling solver on the 4i dataset.
 \algo returns better couplings, in terms of the OT cost and the entropy, for a fraction of \citeauthor{sinkhorn1964relationship} iterations, while still returning a coupling that has the same deviation to the original marginals.
The \textit{(top)} row is computed using $h=\|.\|^2_2$, the \textit{(bottom)} row shows results for the cost $h = \tfrac1p\|\cdot\|^p_p$ where $p=1.5$.}
\vspace{-15pt}
    \label{fig:coupling_performance}
\end{figure}

\subsection{\algo as a Coupling Solver}\label{sec:coupling_experiment}
In this section, we benchmark the ability of \algo to return a coupling, and compare it to that of the  \citeauthor{sinkhorn1964relationship} algorithm. 
Comparing coupling solvers is rife with challenges, as their time performance must be compared comprehensively by taking into account three crucial metrics: \textit{(i)} the cost of transport according to the coupling $\mathbf{P}$, that is, $\langle \mathbf{P}, \mathbf{C}\rangle$, \textit{(ii)} the entropy $E(\mathbf{P})$, and \textit{(iii)} satisfaction of marginal constraints $\|\mathbf{P}\mathbf{1}_m - \ba\|_1 +\|\mathbf{P}^T\mathbf{1}_n - \bb\|_1$.
Due to our threshold schedule $(\tau_k)_k$, as detailed in \cref{sec:implementation}, both approaches are guaranteed to output couplings that satisfy the same threshold for criterion \textit{(iii)}, leaving us only three quantities to monitor: compute effort here quantified as total number of \citeauthor{sinkhorn1964relationship} iterations, summed over all $K$ steps for \algo), transport cost and entropy. While compute effort and transport cost should, ideally, be as small as possible, certain applications requiring, e.g., differentiability~\citep{cuturi2019differentiable} or better sample complexity~\citep{genevay2019sample}, may prefer higher entropies.  

To monitor these three quantities, and cover an interesting space of solutions that, we run \citeauthor{sinkhorn1964relationship}'s algorithm for a logarithmic grid of $\eps=\beta\eps_0$ values (here $\eps_0$ is defined in Line~\ref{lst:eps0} of \Cref{alg:eps_sched}), and compare it to constant-speed \algo with $K = \{2, 4, 8\}$. 
Because one cannot directly compare regularizations, we explore many choices to schedule $\eps$ within \algo. Following the default strategy used in OTT-JAX~\citep{cuturi2022optimal}, we set at every iterate $k$, $\eps_k = \theta \bar c_k$, where $\bar c_k$ is 5\% of the the mean of the cost matrix at that iteration, as detailed in \cref{app:experiment_details}. We do not use \cref{alg:eps_sched} since it returns a regularization schedule that is tuned for map estimation, while the goal here is to recover couplings that are comparable to those outputted by \citeauthor{sinkhorn1964relationship}.
We set the threshold for marginal constraint satisfaction for both algorithms as $\tau_K=\tau=0.001$ and run all algorithms to convergence, with infinite iteration budget.
For the coupling experiments, we use the single-cell multiplex data of \citet{bunne2023}, reflecting morphological features and protein intensities of melanoma tumor cells.
The data describes $d=47$ features for $n\approx 11,000$ control cells, and $m\approx 2,800$ treated cells, for each of 34 drugs, of which we use only 6 at random.
To align the cell populations, we consider two ground costs: the squared-Euclidean norm $\|\cdot\|^2$ 
as well as $h=\tfrac{1}{p}\|\cdot\|_p^p$, with $p=1.5$. \looseness-1

Results for $6$ drugs are displayed in Figure~\ref{fig:coupling_performance}.
The \textit{area} of the marker reflects the total number of \citeauthor{sinkhorn1964relationship} iterations needed for either algorithm to converge to a coupling with a threshold $\tau=10^{-3}$. The values for $\beta$ and $K$ displayed in the legend are encoded using \textit{colors}. The global scaling parameter for \algo is set to $\theta=2^{-4}$.  \cref{fig:coupling_exp_extended_l2}~and~\ref{fig:coupling_exp_extended_l15} visualize other choices for $\theta$.
These results prove that \algo provides a competitive alternative to \citeauthor{sinkhorn1964relationship}, to compute couplings that yield a small entropy and cost at a low computational effort, while satisfying the same level of marginal constraints. 
\looseness-1

\subsection{Scalability of \algo}

\begin{figure}
\centering
\begin{minipage}[t]{0.58\textwidth}
\centering
    \caption{We consider the optimal assignment problem between all CIFAR images and their blurry CIFAR counterparts using the $\ell_2^2$ loss. A small subset of 3 original images on the left can be compared with their blurred counterpart on the right, with $\sigma = 4$. The optimal coupling for this task is the identity, which we compare with couplings recovered by our methods at large scales.    \label{fig:cifar_benchmark} \looseness-1}
\includegraphics[width=0.9\linewidth]{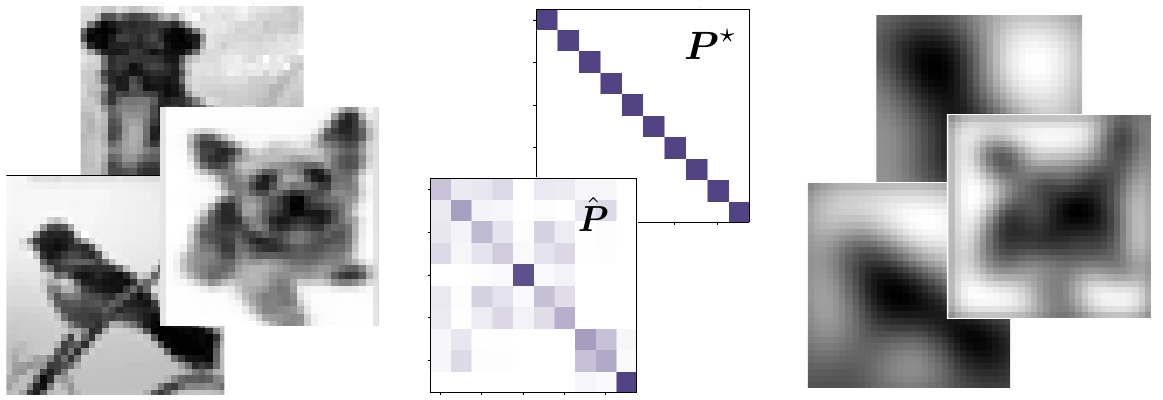}
    \end{minipage}
    \hfill
    \begin{minipage}[t]{0.4\textwidth}
        \centering
        \captionof{table}{Coupling recovery, quantified as trace, and KL divergence from identity matrix, for coupling matrices obtained with \algo and Sinkhorn, and blur strengths $\sigma=2,4$. 
        \algo is run for $K=4$ and with the constant-speed schedule. \label{tab:cifar}\looseness-1}
{\scalebox{.89}{
\begin{tabular}{|c|c|*{2}{c|}}
\hline
 \multicolumn{2}{|c|}{$\sigma$} & 2& 4\\
\hline
 \multirow{2}{*}{Sinkhorn} & $\mathrm{Tr}$ & 
0.9999&0.9954
 \\
 \cline{2-4}
&$\mathrm{KL}$ & 
 0.00008 & 0.02724\\
  \cline{2-4}
&\# iters & 
10&2379
\\
\hline
 \multirow{2}{*}{\algo} & $\mathrm{Tr}$ & 
1.000&
0.9989\\
 \cline{2-4}
&$\mathrm{KL}$ & 
0.00000 & 0.00219\\ 
 \cline{2-4}
& \# iters &
40 &
1590 \\
\hline
\end{tabular}}}
    \end{minipage}
    \vspace{-10pt}
\end{figure}

Real-world experiments on pre-processed single-cell data are often run with limited sample sizes ($n\simeq10^3$) and medium data dimensionality ($d\leq 200$). As a result they are not suitable to benchmark OT solvers at larger scale.
To address this limitation, we design a challenging large-scale (large $n$, large $d$) OT problem on real data, for which the ground-truth is known. We believe our approach can be replicated to create benchmarks for OT solvers.
We consider the entire grayscale CIFAR10 dataset \citep{krizhevsky2009learning} for which $n = 60,000$ and $d = 32\times 32 = 1024$.
We consider the task of matching these $n$ images to their blurred counterparts, using Gaussian blurs of varying width.
To blur an image \smash{$U \in \sR^{N\times N}$} we use the isotropic Gaussian kernel \smash{$K =[\exp\left(-(i-j)^2/(\sigma N^2)\right)]_{ij}$} for $i, j \leq N$ \citep[c.f. Remark 4.17,][]{peyre2019computational}, and define the Gaussian blur operator as \smash{$G(U) \coloneqq KUK \in \sR^{N\times N}$}. 
The crucial observation we make in \cref{prop:cifar} is that, when using the squared Euclidean ground cost $\ell^2_2$, the optimal matching is necessarily equal to the \textit{identity} (i.e. each image must be necessarily matched to its blurred counterpart), as pictured in (\cref{fig:cifar_benchmark}).

\begin{proposition}\label{prop:cifar}
    Let \smash{$\hat\mu = \tfrac1n\sum_{s\leq n} \delta_{U_s}$} be the empirical distribution over $n$ images and define $\hat \nu \coloneqq G_\# \hat \mu$ where $G$ is the Gaussian blur operator with $\sigma<\infty$.
    Then $\mP^\star$ the optimal coupling between $(\hat \mu, \hat \nu)$ with the \smash{$h = \tfrac{1}{2}\norm{\cdot}_2^2$} cost is the normalized $n$-dimensional identity matrix $\Id/n$.
\end{proposition}

In light of \cref{prop:cifar}, we generate two blurred datasets using a Gaussian kernel with $\sigma = 2$ and 
$\sigma = 4$ (see \cref{fig:cifar_noises}). We then use \algo with and \citeauthor{sinkhorn1964relationship}'s Algorithm to match the blurred dataset back to the original CIFAR10 (de-blurring). The hyper-parameter configurations are the same as \cref{sec:coupling_experiment}, with $\beta = \theta = 2^{-4}$.
We evaluate the performance of the OT solvers by checking how close the trace of the recovered coupling $\mathrm{Tr}(\hat \mP)$ is to $1.0$, or with the KL divergence from the ground-truth, that is, $\mathrm{KL}(\mP^\star || \hat\mP) = - \log n - n \sum_{i\leq n} \log(\hat \mP_{ii})$.

\cref{tab:cifar} compares the performance of \algo and \citeauthor{sinkhorn1964relationship}, along with the number of iterations needed to achieve this performance. Both algorithms scale well and show high accuracy, while requiring a similar amount of computation.
We highlight that at this scale, simply storing the cost or coupling matrices would require about 30Gb. The experiment has to happen across multiple GPUs. Thanks to its integration in JAX and OTT-JAX \citep{ott-jax}, \algo supports sharding by simply changing a few lines of code. The algorithms scales seamlessly and each run takes about 15 minutes, on a single node of 8 A100 GPUs.
This experiment sets a convincing example on how \algo scales to much larger (in $n$ and $d$) problems than considered previously.

\section*{Conclusion}\label{sec:concl}
In this work, we proposed \algo, a new family of EOT solvers that blend dynamic and static formulations of OT by using the \citeauthor{sinkhorn1964relationship} algorithm as a subroutine within a progressive scheme. 
\algo aims to provide practitioners with an alternative to the \citeauthor{sinkhorn1964relationship} algorithm that \textit{(i)} does not fail when instantiated with uninformed or ill-informed $\eps$ regularization, thanks to its self-correcting behavior and our simple $\eps$-scheduling scheme that is informed by the dispersion of the \emph{target distribution},  
\textit{(ii)} performs at least as fast as \citeauthor{sinkhorn1964relationship} when used to compute couplings between point clouds, and \textit{(iii)} provides a reliable out-of-the-box OT map estimator that comes with a non-asymptotic convergence guarantee. 
We believe \algo can be used as a strong baseline to estimate  \citeauthor{Monge1781} maps.\looseness-1

\bibliographystyle{abbrvnat}
\bibliography{refs}
\newpage
\appendix
\section{Additional Experiments} \label{app:experiments_additional}

In \cref{sec:map_experiment}, we demonstrated the performance of \algo for map estimation on the sci-Plex dataset. Here, we present a similar experiment on the 4i data, and extend all map experiments to the case of a general translation invariant cost function, $h = 1.5\norm{\cdot}_{1.5}$.
In \cref{tab:cell_benchmarks_15norm} and \cref{tab:cell_benchmarks_4i} we show $D_{\eps_D}((\Tprog)_\#\mX, \mY; h)$ the sinkhorn divergence using the cost function $h$.

\begin{table}[h]
    \centering
    \caption{Performance of algorithms on sci-Plex data, w.r.t $D_{\eps_D}((\Tprog)_\#\mX, \mY; h)$ with the 1.5-norm cost. Reported numbers are the average of 5 runs, together with the standard error.\looseness-1}
    \label{tab:cell_benchmarks_15norm}
{\scalebox{.82}{\begin{tabular}{|c|*{9}{c|}c|}
\hline
Drug & \multicolumn{3}{c|}{Belinostat} & \multicolumn{3}{c|}{Givinostat} & \multicolumn{3}{c|}{Hesperadin} & \multirow{2}{*}{\makecell{ 5-drug\\rank}}\\
\cline{1-10}
$d_\mathrm{PCA}$& 16 & 64 & 256 & 16 & 64 & 256 & 16 & 64 & 256& \\
\hline
\algo &
\textbf{2.03}$\pm$0.02&
\textbf{7.11}$\pm$0.03&
\textbf{18.6}$\pm$0.1&
\textbf{2.04}$\pm$0.07&
7.1$\pm$0.1&
19.5$\pm$0.1&
\textbf{2.4}$\pm$0.1&
\textbf{7.7}$\pm$0.1&
\textbf{20.2}$\pm$0.4&
\textbf{1}\\
\hline
EOT & 
2.07$\pm$0.02&
7.22$\pm$0.06&
18.8$\pm$0.2&
2.0$\pm$0.1&
\textbf{7.1}$\pm$0.1&
\textbf{19.5}$\pm$0.1&
2.6$\pm$0.2&
8.1$\pm$0.2&
20.6$\pm$0.6
&2\\
\hline
\makecell{Debiased EOT} & 
3.90$\pm$0.04&
13.8$\pm$0.1&
37.6$\pm$0.2&
4.2$\pm$0.1&
14.4$\pm$0.1&
38.7$\pm$0.2&
3.6$\pm$0.2&
13.0$\pm$0.2&
36.0$\pm$0.5&
4\\
\hline
\makecell{Monge Gap} &
2.4$\pm$0.1&
8.6$\pm$0.1&
34.2$\pm$0.3 & 
2.3$\pm$0.1&
8.5$\pm$0.2&
34.8$\pm$0.3&
3.7$\pm$0.5&
10.4$\pm$0.5&
36.0$\pm$0.8&
3\\
\hline
\end{tabular}}}
\end{table}

\begin{table}[h]
    \centering
    \caption{Performance of algorithms on 4i data, w.r.t $D_{\eps_D}((\Tprog)_\#\mX, \mY; h)$ where $h$ is reported in the table. Reported numbers are the average of 10 runs, together with the standard error.\looseness-1}
    \label{tab:cell_benchmarks_4i}
{\scalebox{.84}{\begin{tabular}{|c|*{3}{c|}*{3}{c|c|}}
\hline
Drug/ & cisplatin & dasatinib & everolimus & vindesine & staurosporine & decitabine & overall\\
Cost & $\ell_2$ & $\ell_2$ & $\ell_2$ & 1.5$\norm{\cdot}_{1.5}$ & 1.5$\norm{\cdot}_{1.5}$ & 1.5$\norm{\cdot}_{1.5}$ & rank\\
\hline
\algo & \textbf{1.68}$\pm$0.04&
2.7$\pm$0.1&
\textbf{1.66}$\pm$0.06&
2.21$\pm$0.03&
2.74$\pm$0.08&
1.66$\pm$0.01 & 2\\
\hline
EOT  & 2.72$\pm$0.05&
2.17$\pm$0.03&
4.86$\pm$0.11&
2.91$\pm$0.09&
1.96$\pm$0.04&
1.73$\pm$0.04 & 3=\\
\hline
\makecell{Debiased EOT}  & 1.79$\pm$0.07&
\textbf{1.65}$\pm$0.05&
2.98$\pm$0.29&
2.86$\pm$0.25&
\textbf{1.81}$\pm$0.05&
\textbf{1.64}$\pm$0.05 & \textbf{1}\\
\hline
\makecell{Monge Gap}  & 1.7$\pm$0.1&
3.3$\pm$0.2&
1.7$\pm$0.1&
\textbf{2.17}$\pm$0.05&
3.03$\pm$0.08&
1.81$\pm$0.05 & 3=\\
\hline
ICNN  & 1.74$\pm$0.08&
3.8$\pm$0.7&
1.88$\pm$0.05& - & - & - & - \\
\hline
\end{tabular}}}
\end{table}

\textbf{GMM Benchmark.} To provide further evidence on performance of \algo, we benchmark the map estimation methods on high-dimensional Gaussian Mixture data, using the dataset of \citet{korotin2021neural} for $d=128, 256$. In this experiment the ground truth maps are known and allow us to compare the algorithms more rigorously using the empirical $\ell_2$ distance of the maps, as defined in section \cref{sec:map_experiment}. Shown in \cref{tab:gmm_benchmark}, we consider $n_{\mathrm{test}}=500$ test points and $n_{\mathrm{train}} = 8000$ and $9000$ training points, respectively for each dimension. We have also included a variant of the entropic estimator which uses the default value of the OTT-JAX library for $\varepsilon$, and unlike other EOT algorithms, is not cross-validated. This is to demonstrate the significant effect that $\varepsilon$ has on \citeauthor{sinkhorn1964relationship} solvers.

\begin{table}[h]
  \centering
    \caption{GMM benchmark. The Table shows the MSE, average of $\|\hat{\vy}-\vy_{\mathrm{test}}\|^2_2$ for $n_{\mathrm{test}} = 500$ points, where $\hat\vy = \hat T(\vx_{\mathrm{test}})$ and the ground truth is $\vy_{\mathrm{test}}$.\label{tab:gmm_benchmark}}
\begin{tabular}{|c|c|c|}
\hline
& $d=128$ & $d = 256$  \\
\hline
\algo & \textbf{0.099}$\pm$0.009& \textbf{0.12}$\pm$0.01\\
\hline
EOT & 0.12$\pm$0.01& 0.16$\pm$0.02 \\
\hline
Debiased EOT& 0.11$\pm$0.01& 0.128$\pm$0.002\\
\hline
Untuned EOT& 0.250$\pm$0.023&
0.276$\pm$0.006\\
\hline
Monge Gap &0.36$\pm$0.02&0.273$\pm$0.005 \\
\hline
ICNN & 0.177$\pm$0.023&
0.117$\pm$0.005\\
\hline
\end{tabular}
\end{table}

\begin{figure}[t]
    \centering
    \includegraphics[width=0.6\linewidth]{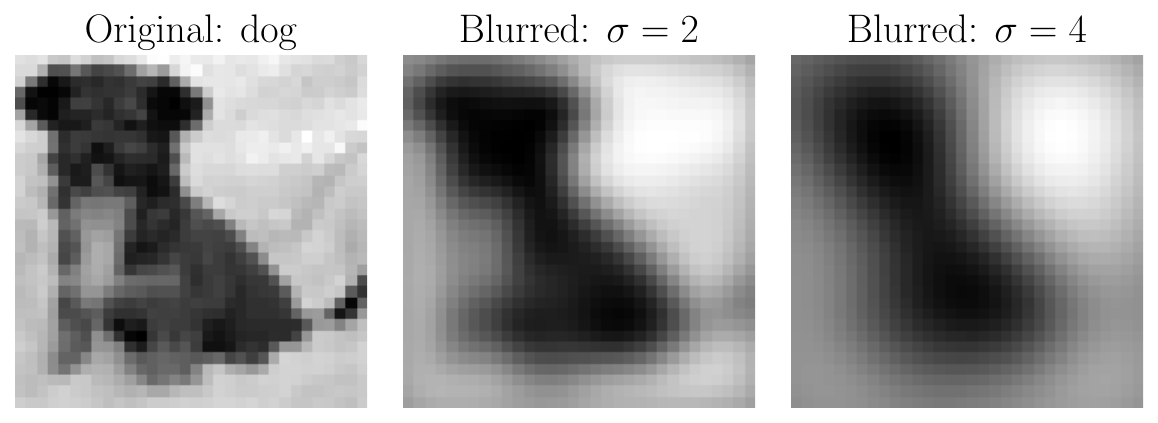}
    \caption{Example of a CIFAR10 image and blurred variant. We match blurry images to the originals.\looseness-1}
    \label{fig:cifar_noises}
\end{figure}

\section{Details of Experiments} \label{app:experiment_details}
\textbf{Generation of \cref{fig:figure1} and \cref{fig:couplings}.}
We consider a toy example where the target and source clouds are as shown in \cref{fig:figure1}.
We visualize the entropic map \citep{pooladian2021entropic}, its debiased variant \citep{pooladian2022debiaser} and \algo, where we consider a decelerated schedule with $6$ steps, and only visualize steps $k=3, 5$ to avoid clutter.
The hyperparameters of the algorithms are set as described in \cref{sec:experiments}.
\cref{fig:couplings} shows the coupling matrix corresponding to the same data, resulting from the same solvers.

\textbf{Sinkhorn Divergence and its Regularization Parameter.}
In some of the experiments, we calculate the Sinkhorn divergence between two point clouds as a measure of distance.
In all experiments we set the value of $\eps_D$ and according to the geometry of the target point cloud. 
In particular, we set $\eps$ to be default value of the OTT-JAX \cite{ott-jax} library for this point cloud via \texttt{ott.geometry.pointcloud.PointCloud(Y).epsilon}, that is, $5\%$ of the average cost matrix, within the target points.

\textbf{Details of Scheduling $(\alpha_k)_k$.} \cref{tab:alpha_sched} specifies our choices of $\alpha_k$ for the three schedules detailed in \cref{sec:implementation}.
\cref{fig:lin_vs_exp_progot} compares the performance of \algo with constant-speed schedule in red, with the decelerated (Dec.) schedules in green. The figure shows results on the sci-Plex data (averaged across 5 drugs) and the Gaussian Mixture synthetic data.
We observe that the algorithms perform roughly on par, implying that in practice \algo is robust to choice of $\alpha$.

\begin{table}[h]
    \centering
      \caption{Scheduling Schemes for $\alpha_k$.}
    \begin{tabular}{c|c|c}
    Schedule & $\alpha_k = \alpha(k)$ &  $t_k=t(k)$\\
    \hline
      Decelerated   &  $1/e$ & $\frac{1-e^{-(k+1)}}{e-1}$\\
      Constant-Speed & $\frac{1}{K-k+2}$ & $\frac{k}{K}$\\
      Accelerated & $\frac{2k-1}{(K+1)^2-(k-1)^2}$& $\left( \frac{k}{K}\right)^2$\\
    \end{tabular}
    \label{tab:alpha_sched}
\end{table}

\textbf{Details of Scheduling $(\eps_k)_k$.}
For map experiments on the sci-Plex data \citep{srivatsan2020massively}, we schedule the regularization parameters via \cref{alg:eps_sched}. We set $\beta_0 = 5$ and consider the set $s_p = \{ 0.1, 0.5, 1, 5, 10, 20\}$.
For coupling experiments on the 4i data \citep{gut2018} we set the regularizations as follows.
Let \smash{$\vx_1\ith{k}, \dots, \vx_n\ith{k}$} denote the interpolated point cloud at iterate $k$ (according to Line 7, \cref{alg:progot}) and recall that $\vy_1, \dots, \vy_m$ is the target point cloud.
The scaled average cost at this iterate is \smash{$\bar c_k = \sum_{i,j}h(\vx\ith{k}_i - \vy_j)/(20mn)$}, which is the default value of $\varepsilon$ typically used for running \citeauthor{sinkhorn1964relationship}.
Then for every $k \in [K]$, we set $\eps_k = \theta \bar c_k$ to make \algo compatible to the $\beta$ regularization levels of the bench-marked \citeauthor{sinkhorn1964relationship} algorithms. For \cref{fig:coupling_performance}, we have set $\theta = 2^{-4}$. In \cref{fig:coupling_exp_extended_l2} and \cref{fig:coupling_exp_extended_l15}, we visualize the results for $\theta \in \{2^{-7}, 2^{-4}, 2^{-1}\}$ to give an overview of the results using small and larger scaling values.

\textbf{Compute Resources.} Experiments were run on a single Nvidia A100 GPU for a total of 24 hours. Smaller experiments and debugging was performed on a single MacBook M2 Max.

\begin{figure}
    \centering
    \includegraphics[width = \textwidth]{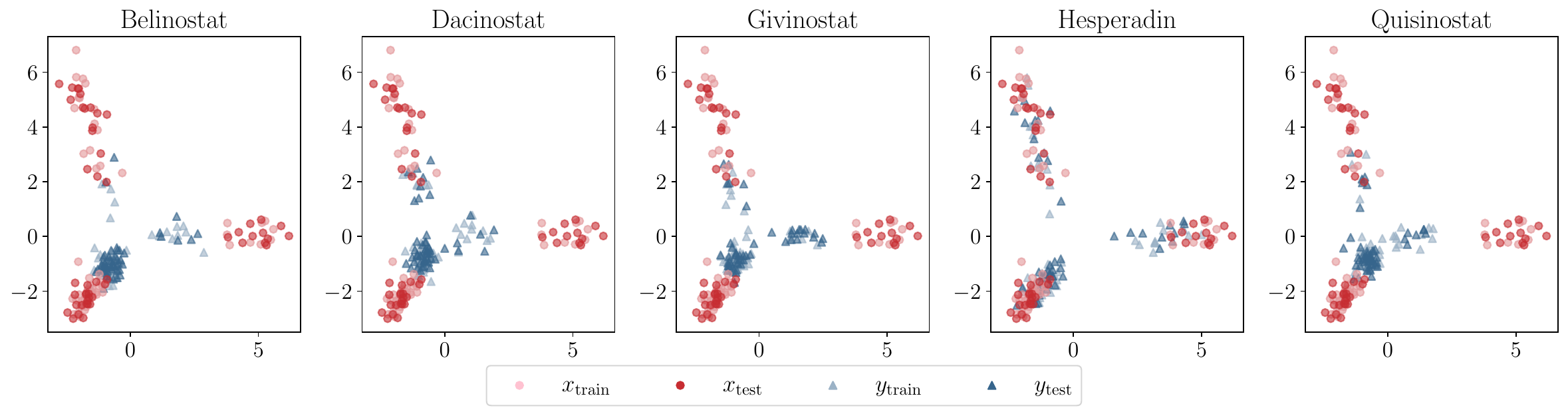}
    \caption{Overview of the single cell dataset \citet{srivatsan2020massively}. We show the first two PCA dimensions performed on the training data, and limit the figure to 50 samples. The point cloud $(\vx_{\mathrm{train}}, \vx_{\mathrm{test}})$ shows the control cells and $(\vy_{\mathrm{train}}, \vy_{\mathrm{test}})$ are the perturbed cells using a specific drug.\looseness-1}
    \label{fig:scrna_distrib}
\end{figure}

\begin{figure}
    \centering
    \includegraphics[width = \textwidth]{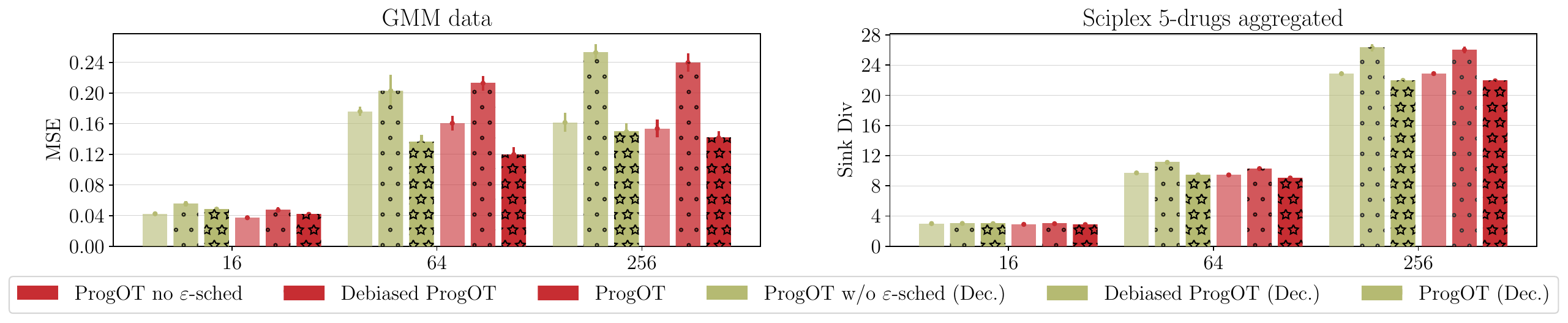}
    \caption{Comparison of constant speed vs decelerated \algo.}
    \label{fig:lin_vs_exp_progot}
\end{figure}

\begin{figure}[h]
    \centering
    \includegraphics[width = \textwidth]{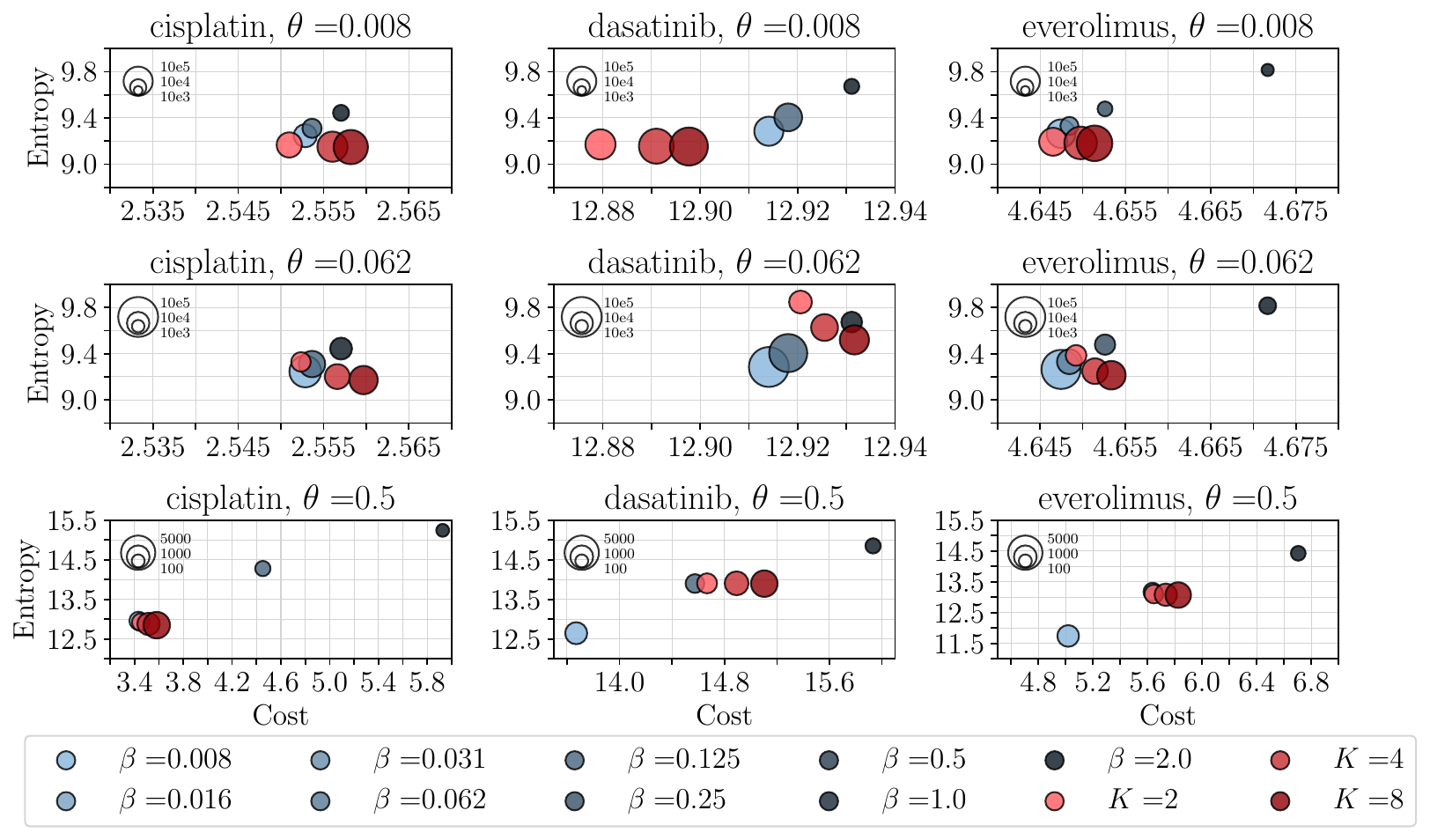}
    \caption{Comparison of \algo and Sinkhorn as coupling solvers for $h(\cdot) = \|\cdot\|_2^2$ on the 4i dataset. Rows show different choices of regularization $\theta$ for \algo as detailed in \cref{app:experiment_details}.}
    \label{fig:coupling_exp_extended_l2}
\end{figure}
\begin{figure}
    \centering
    \includegraphics[width = \textwidth]{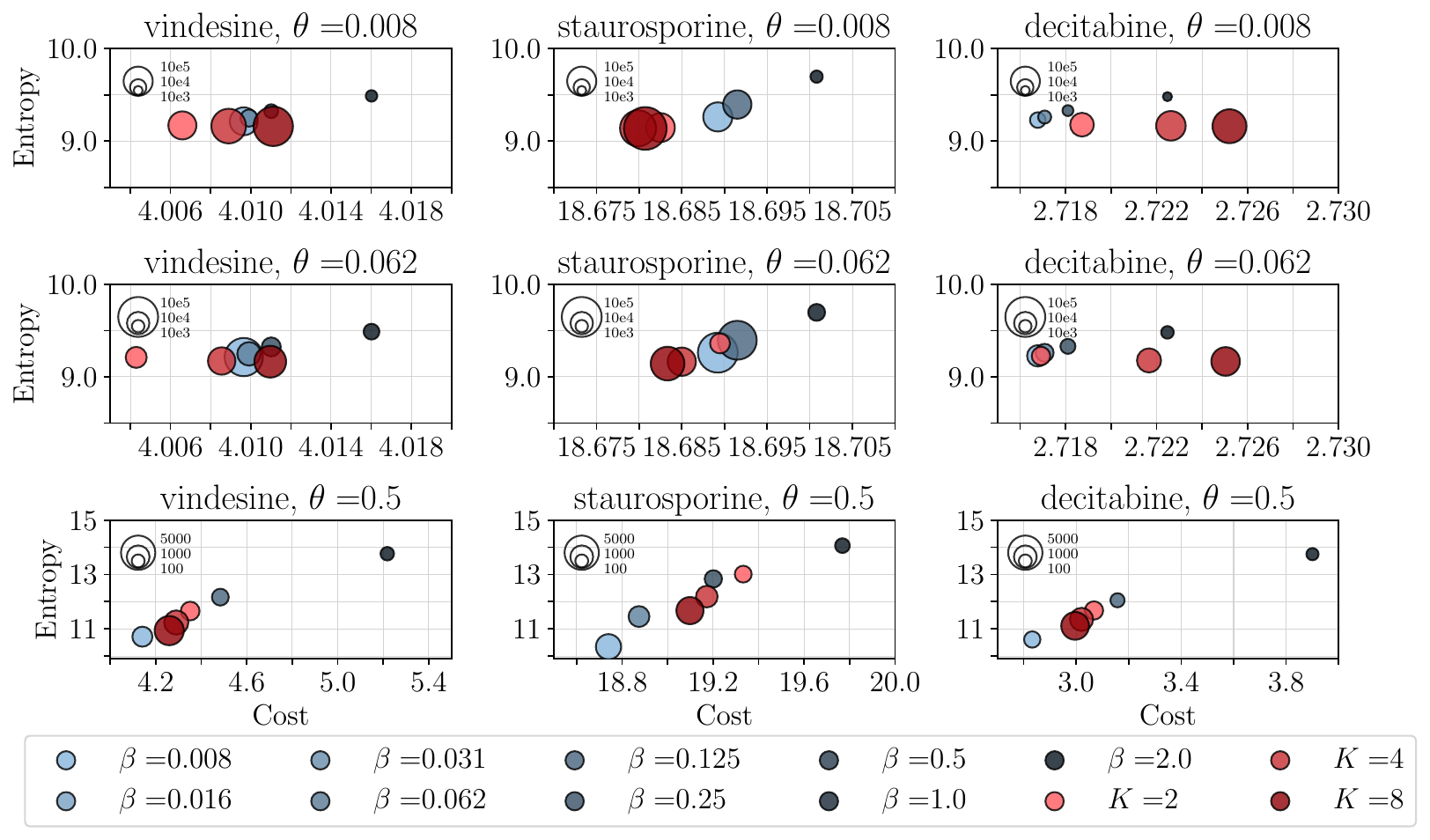}
    \caption{Comparison of \algo and Sinkhorn as coupling solvers for $h(\cdot) = \|\cdot\|_{1.5}^{1.5}/1.5$ on the 4i dataset. Rows show different choices of regularization $\theta$ for \algo as detailed in \cref{app:experiment_details}.}
    \label{fig:coupling_exp_extended_l15}
\end{figure}
\begin{figure}
    \centering
    \includegraphics[width = \textwidth]{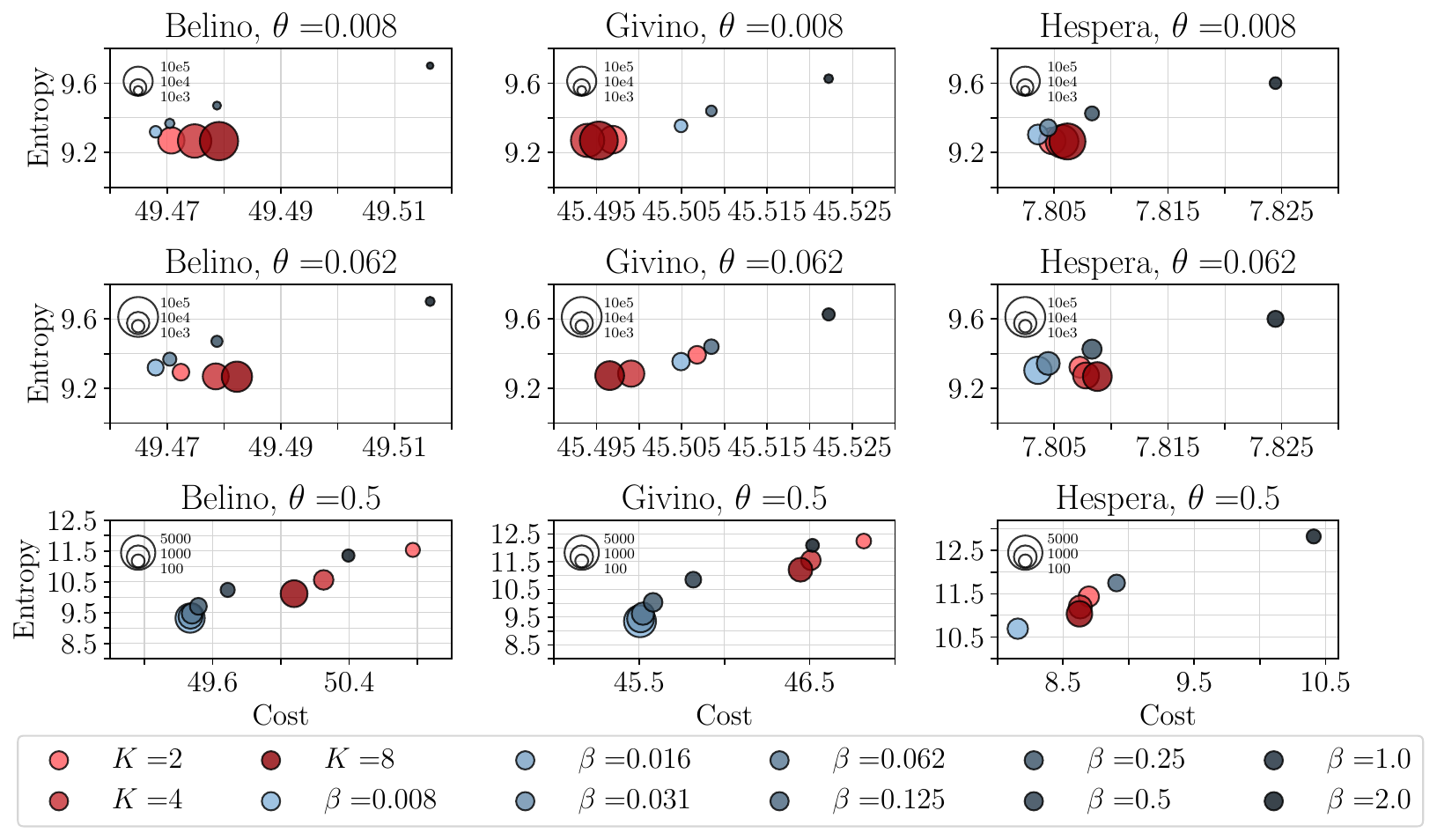}
    \caption{Comparison of \algo and Sinkhorn as coupling solvers for $h(\cdot) = \|\cdot\|_2^2$ on the sci-Plex dataset. Rows show different choices of regularization $\theta$ for \algo as detailed in \cref{app:experiment_details}.}
    \label{fig:sciplex_extended_l2}
\end{figure}
\begin{figure}
    \centering
    \includegraphics[width = \textwidth]{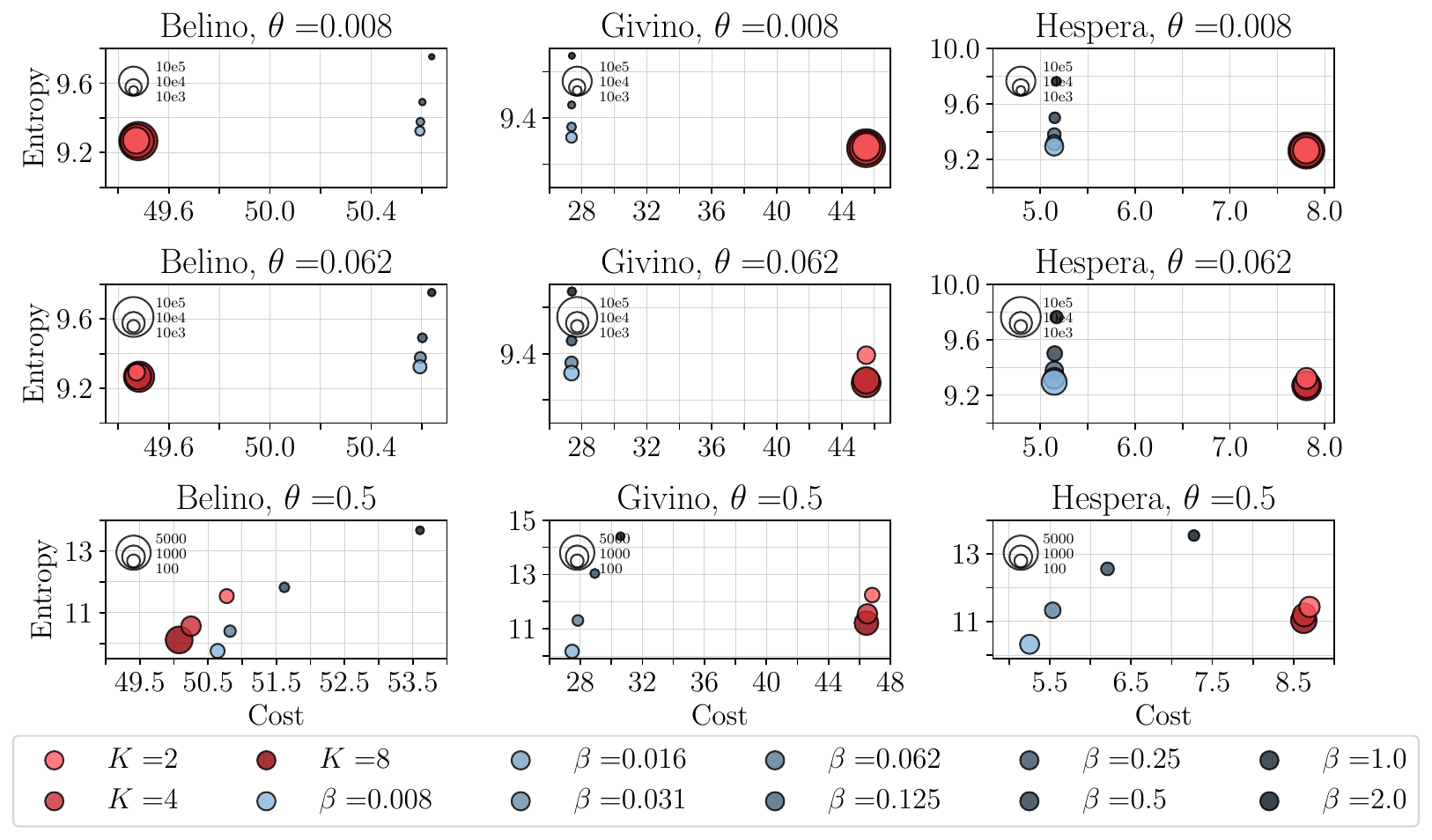}
    \caption{Comparison of \algo and Sinkhorn as coupling solvers for $h(\cdot) = \|\cdot\|_{1.5}^{1.5}/1.5$ on the sci-Plex dataset. Rows show different choices of regularization $\theta$ for \algo as detailed in \cref{app:experiment_details}.}
    \label{fig:sciplex_extended_l15_1}
\end{figure}
\begin{figure}
    \centering
    \includegraphics[width = \textwidth]{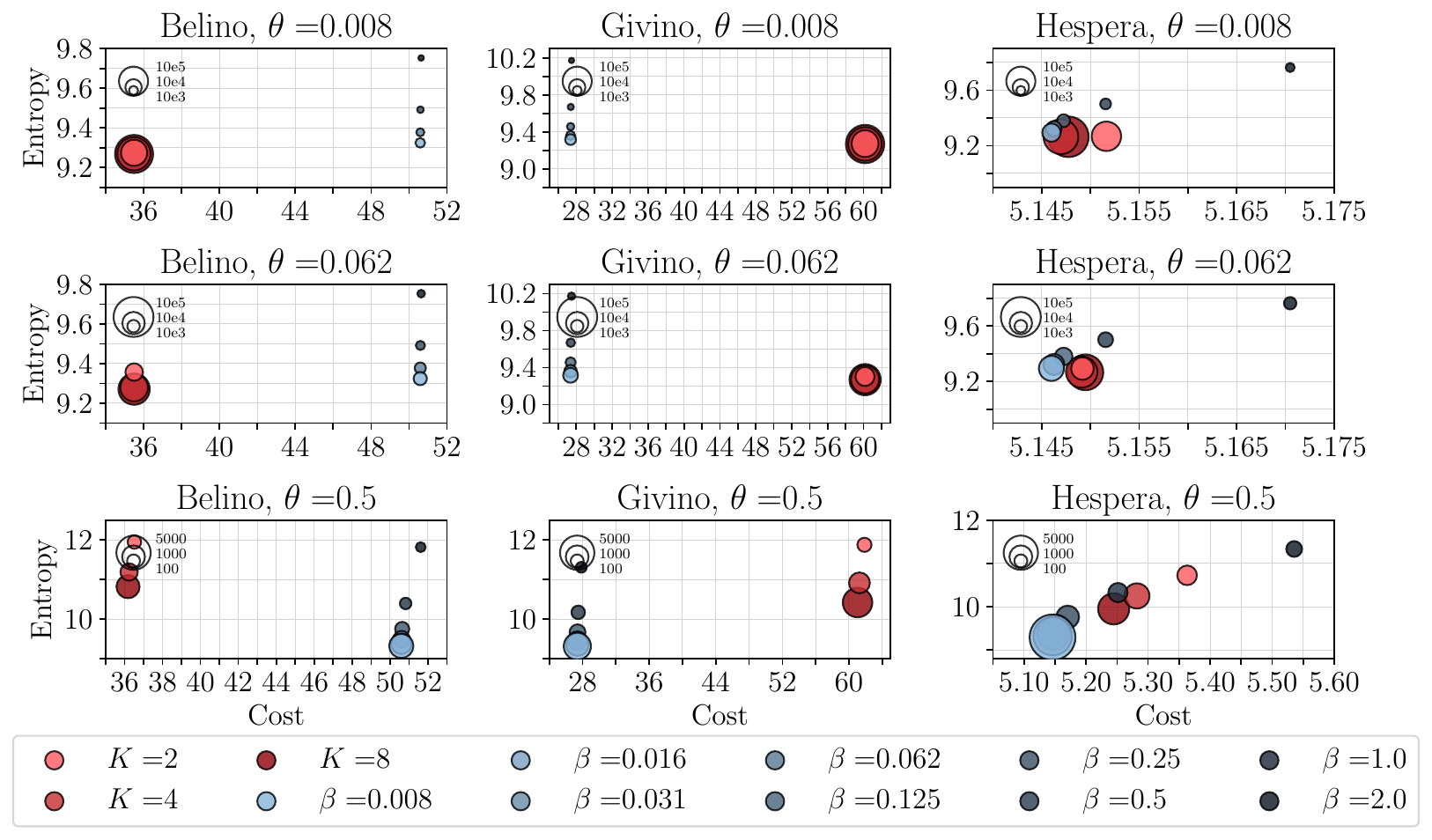}
    \caption{Comparison of \algo and Sinkhorn as coupling solvers for $h(\cdot) = \|\cdot\|_{1.5}^{1.5}/1.5$ on the sci-Plex dataset. Rows show different choices of regularization $\theta$ for \algo as detailed in \cref{app:experiment_details}. The threshold for marginals is also scheduled here, starting from $\tau = 0.01$ and reaching $\tau = 0.001$ to match Sinkhorn.}
    \label{fig:sciplex_extended_l15_2}
\end{figure}

\section{Proofs}\label{app:proofs}
\paragraph{Preliminaries.} Before proceeding with the proofs, we collect some basic definitions and facts. First, we write the the $p$-Wasserstein distance for any $p \geq 1$:
\begin{align*}
    W_p(\mu,\nu) \defeq \Bigl(\inf_{\pi \in \Pi(\mu,\nu)} \iint \|x-y\|^p \dd \pi(x,y)\Bigr)^{1/p}\,.
\end{align*}
Moreover, it is well-known that $p$-Wasserstein distances are ordered for $p \geq 1$: for $1 \leq p \leq q$, it holds that $W_p(\mu,\nu) \leq W_q(\mu,\nu)$  \citep[cf. Remark 6.6,][]{villani2009optimal}. 

For the special case of the $1$-Wasserstein distance, we have the following dual formulation
\begin{align*}
    W_1(\mu,\nu) = \sup_{f \in \text{Lip}_1} \int f \dd(\mu-\nu)\,,
\end{align*}
where $\text{Lip}_1$ is the space of $1$-Lipschitz functions \citep[cf. Theorem 5.10,][]{villani2009optimal}. 

Returning to the $2$-Wasserstein distance, we will repeatedly use the following two properties of optimal transport maps. First, for any two measures $\mu,\nu$ and an $L$-Lipschitz map $T$, it holds that
\begin{align}\label{eq:pushingsame}
    W_2(T_\#\mu,T_\#\nu) \leq LW_2(\mu,\nu)\,.
\end{align}
This follows from a coupling argument. In a similar vein, we will use the following upper bound on the Wasserstein distance between the pushforward of a source measure $\mu$ by two different optimal transport maps $T_a$ and $T_b$:
\begin{align}\label{eq:pushingaway}
    W_2^2((T_a)_\#\mu, (T_b)_\#\mu) \leq \|T_a - T_b\|^2_{L^2(\mu)}\,,
\end{align}

\paragraph{Notation conventions.} For an integer $K \in \mathbb{N}$, $[K]\defeq \{0,\ldots, K\}$. We write $a \lesssim b$ to mean that there exists a constant $C > 0$ such that $a \leq C b$. A constant can depend on any of the quantities present in \textbf{(A1)} to \textbf{(A3)}, as well as the support of the measures, and the number of iterations in \cref{alg:progot}.
The notation $a \lesssim_{\log(n)} b$ means that $a \leq C_1 (\log n)^{C_2} b$ for positive constants $C_1$ and $C_2$.

\subsection{Properties of entropic maps}\label{app:entmap_properties}
Before proving properties of the entropic map, we first recall the generalized form of \eqref{eq:entdual}, which holds for arbitrary measures (cf. \citet{genevay2019entropy}):
\begin{align}\label{eq:entwassdist_dual_appendix}
    \sup_{(f,g)\in L^1(\mu)\times L^1(\nu)} \int f \dd \mu + \int g \dd\nu - \eps \iint e^{(f(x)+g(y) - \tfrac12\|x-y\|^2)/\eps}\dd\mu(x)\dd\nu(y) + \eps\,.
\end{align}
When the entropic dual formulation admits maximixers, we denote them by $(f_\eps,g_\eps)$ and refer to them as \emph{optimal entropic Kantorovich} potentials \citep[e.g.,][Theorem 7]{genevay2019entropy}. Such potentials always exist if $\mu$ and $\nu$ have compact support.

We can express an entropic approximation to the optimal transport coupling $\pi_0$ as a function of the dual maximizers \citep{Csi75}:
\begin{align}\label{eq:pieps_star}
    \pi_\eps(x,y)\defeq \gamma_{\eps}(x,y) \dd\mu(x) \dd \nu(y) \defeq \exp\Bigl(\frac{f_\eps(x) + g_\eps(y) - \tfrac12\|x-y\|^2}{\eps}\Bigr)\dd\mu(x) \dd \nu(y)\,. 
\end{align}
When necessary, we will be explicit about the measures that give rise to the entropic coupling. For example, in place of the above, we would write
\begin{align}\label{eq:gamma_appendix}
    \pi_\eps^{\mu\to\nu}(x,y) = \gamma_\eps^{\mu\to\nu}(x,y)\dd\mu(x)\dd\nu(y)\,.
\end{align}
The population counterpart to \eqref{eq:Tent}, the entropic map from $\mu$ to $\nu$, is then expressed as
\begin{align*}%
    T_\eps^{\mu\to\nu}(x) \defeq \E_{\pi_\eps}[Y|X=x]\,,
\end{align*}
and similarly the entropic map from $\nu$ to $\mu$ is
\begin{align*}%
    T_\eps^{\nu\to\mu}(y) \defeq \E_{\pi_\eps}[X|Y=y]\,.
\end{align*}
We write the forward (resp. backward) entropic Brenier potentials as $\varphi_\eps\defeq \tfrac12\|\cdot\|^2 - f_\eps$ (resp. $\psi_\eps \defeq \tfrac12\|\cdot\|^2 - g_\eps$). By dominated convergence, one can verify that
\begin{align*}
    \nabla\varphi_\eps(x) = T^{\mu\to\nu}_\eps(x)\,, \quad \nabla\psi_\eps(y) = T^{\nu\to\mu}_\eps(y)\,. 
\end{align*}

We now collect some general properties of the entropic map, which we state over the ball but can be readily generalized.
\begin{lemma}\label{lem:lipsch_bound}
Let $\mu, \nu$ be probability measures over $B(0;R)$ in $\R^d$. Then for a fixed $\eps > 0$, it holds that both $T^{\mu\to\nu}_\eps$ and $T^{\nu\to\mu}_\eps$ are Lipschitz with constant upper-bounded by $R^2/\eps$.
\end{lemma}
\begin{proof}[Proof of \cref{lem:lipsch_bound}]
We prove only the case for $T^{\mu\to\nu}_\eps$ as the proof for the other map is completely analogous. It is well-known that the Jacobian of the map is a symmetric positive semi-definite matrix: $\nabla T_\eps^{\mu\to\nu}(x) = \eps^{-1}\text{Cov}_{\pi_\eps}(Y|X=x)$ (see e.g., \citet[Lemma 1]{chewi2023entropic}). Since the probability measures are supported in a ball of radius $R$, it holds that $\sup_x \|\text{Cov}_{\pi_\eps}(Y|X=x)\|_{\text{op}} \leq R^2$, which completes the claim. 
\end{proof}

We also require the following results from \citet{divol2024tight}, as well as the following object: for three measures $\rho,\mu,\nu$ with finite second moments, write
\begin{align*}
    \bar{I} \defeq \iiint \log\Bigl( \frac{\gamma^{\rho\to\mu}_\eps(x,y)}{\gamma^{\rho\to\nu}_\eps(x,z)}\Bigr) \gamma^{\rho\to\mu}_\eps(x,y) \dd\pi(y,z)\dd\rho(x)\,,
\end{align*}
where $\pi$ is an optimal transport coupling for the $2$-Wasserstein distance between $\mu$ and $\nu$, and $\gamma_\eps$ is the density defined in \eqref{eq:gamma_appendix}.

\begin{lemma}\citep[][Proposition 3.7 and Proposition 3.8]{divol2024tight}\label{lem:divol_helper}
    Suppose $\rho,\mu,\nu$ have finite second moments, then
    \begin{align*}
        \eps \bar{I} \leq \iint \langle T_\eps^{\mu\to\rho}(y) - T_\eps^{\nu\to\rho}(z),y-z \rangle \dd\pi(y, z)\,,
    \end{align*}
    and 
    \begin{align*}
        \iint \|T_\eps^{\mu\to\rho}(y) - T_\eps^{\nu\to\rho}(z)\|^2 \dd\pi(y, z) \leq 2 \bar{I} \sup_{v \in \R^d}\| \mathrm{Cov}_{\pi_\eps^{\rho\to\nu}}(X|Y=v)\|_{\mathrm{op}} \,.
    \end{align*}   
\end{lemma}

We are now ready to prove \cref{prop:stability_phi}. We briefly note that stability of entropic maps and couplings has been investigated by many \citep[e.g.,][]{ghosal2022stability,eckstein2022quantitative,carlier2024displacement}. These works either present \emph{qualitative} notions of stability, or give bounds that depend exponentially on $1/\eps$. In contrast, the recent work of \citet{divol2024tight} proves that the entropic maps are Lipschitz with respect to variations of the target measure, where the underlying constant is linear in $1/\eps$. We show that their result also encompasses variations in the source measure, which is of independent interest.

\begin{proof}[Proof of \cref{prop:stability_phi}]
Let $\pi \in \Gamma(\mu,\mu')$ be the optimal transport coupling from $\mu$ to $\mu'$. By disintegrating and applying the triangle inequality, we have
\begin{align*}
    \int \|T_\eps^{\mu\to\rho}(x) - T_\eps^{\mu'\to\rho}(x)\|\dd \mu(x) & = \iint \|T_\eps^{\mu\to\rho}(x) - T_\eps^{\mu'\to\rho}(x)\|\dd \pi(x, x') \\
    &\leq \iint \|T_\eps^{\mu\to\rho}(x) - T_\eps^{\mu'\to\rho}(x')\| \dd \pi(x, x') \\
    &\qquad \qquad + \|T_\eps^{\mu'\to\rho}(x) - T_\eps^{\mu'\to\rho}(x')\| \dd \pi(x, x') \\
    &\leq \iint \|T_\eps^{\mu\to\rho}(x) - T_\eps^{\mu'\to\rho}(x')\| \dd \pi(x, x') \\
    &\qquad \qquad + \frac{R^2}{\eps}\iint \|x - x'\| \dd \pi(x, x')\\ 
    &\leq \iint \|T_\eps^{\mu\to\rho}(x) - T_\eps^{\mu'\to\rho}(x')\| \dd \pi(x, x') + \frac{R^2}{\eps} W_2(\mu,\mu')\,,
\end{align*}
where the penultimate inequality follows from \cref{lem:lipsch_bound}, and the last step is due to Jensen's inequality. To bound the remaining term, recall that
\begin{align*}
    \sup_{v \in \R^d}\| \text{Cov}_{\pi_\eps^{\rho\to\nu}}(X|Y=v)\|_{\text{op}} \leq R^2\,,
\end{align*}
and by the two inequalities in \cref{lem:divol_helper}, we have (replacing $\nu$ with $\mu'$)
\begin{align*}
    \iint \|T_\eps^{\mu\to\rho}(x) - T_\eps^{\mu'\to\rho}(x')\|^2  \dd \pi(x, x') &\leq 2 \bar{I} R^2 \\
    &= \frac{2R^2}{\eps} (\eps \bar{I}) \\
    &\leq \frac{2R^2}{\eps} \iint \langle T_\eps^{\mu\to\rho}(y) - T_\eps^{\mu'\to\rho}(z),y-z \rangle \dd\pi(y, z) \\
    &\leq \frac{2R^2}{\eps} \Bigl(\iint \|T_\eps^{\mu\to\rho}(x) - T_\eps^{\mu'\to\rho}(x')\|  \dd \pi(x, x')\Bigr) W_2(\mu,\mu')\,,
\end{align*}
where we used Cauchy-Schwarz in the last line. An application of Jensen's inequality and rearranging results in the bound:
\begin{align*}
    \iint \|T_\eps^{\mu\to\rho}(x) - T_\eps^{\mu'\to\rho}(x')\| \dd \pi(x, x') \leq \frac{2R^2}{\eps}W_2(\mu,\mu')\,,
\end{align*}
which completes the claim.
\end{proof}

Finally, we require the following results from \citet{pooladian2021entropic}, which we restate for convenience but under our assumptions.
\begin{lemma}\citep[Two-sample bound:][Theorem 3]{pooladian2021entropic}
    \label{lem:two_sample}
    Consider i.i.d.~samples of size $n$ from each distribution $\mu$ and $\nu$, resulting in the empirical measures $\hat{\mu}$ and $\hat{\nu}$, with the corresponding. Let $\Tent$ be the entropic map between $\hat\mu$ and $\hat\nu$. Under \textbf{(A1)}-\textbf{(A3)}, it holds that
    \[
    \Exp\, \norm{\Tent - T_0}_{L^2(\mu)}^2 \lesssim_{\log(n)} \varepsilon^{-d/2}n^{-1/2} + \varepsilon^2\,. %
    \]
Moreover, if $\eps=\eps(n) \asymp n^{-1/(d+4)}$, then the overall rate of convergence is $n^{-2/(d+4)}$. 
\end{lemma}

\begin{lemma}\citep[One-sample bound:][Theorem 4]{pooladian2021entropic}\label{lem:one_sample}
Consider i.i.d.~samples of size $n$ from $\nu$, resulting in the empirical measure $\hat{\nu}$, with full access to a probability measure $\mu$. Let $\hat{R}_\eps$ be the entropic map from $\mu$ to $\hat\nu$. Under \textbf{(A1)}-\textbf{(A3)}, it holds that
    \[
    \Exp\, \norm{\hat{R}_\eps - T_0}_{L^2(\mu)}^2 \lesssim_{\log(n)} \varepsilon^{1-d/2}n^{-1/2} + \varepsilon^2\,. %
    \]
Moreover, if $\eps=\eps(n) \asymp n^{-1/(d+2)}$, then the overall rate of convergence is $n^{-2/(d+2)}$. 
\end{lemma}

\subsection{Remaining ingredients for the proof of Theorem \ref{thm:consistency_multistep}} \label{app:single_step_bound}
We start by analyzing our Progressive OT map estimator between the iterates. We will recurse on these steps, and aggregate the total error at the end. We introduce some concepts and shorthand notations. \looseness -1

First, the ideal progressive Monge problem: Let $T^{(0)}$ be the optimal transport map from $\mu$ to $\nu$, and write $\mu^{(0)} \defeq \mu$. Then write
\begin{align*}
    S^{(0)}\defeq (1-\alpha_0)\Id  + \alpha_0 T^{(0)}\,,
\end{align*}
and consequently $\mu^{(1)} \defeq (S^{(0)})_\#\mu^{(0)}$.
We can iteratively define $T^{(i)}$ to be the optimal transport map from $\mu^{(i)}$ to $\nu$, and consequently 
\begin{align*}
    S^{(i)} \defeq (1-\alpha_i)\Id  + \alpha_i T^{(i)}\,,
\end{align*}
and thus $\mu^{(i+1)} \defeq (S^{(i)})_\#\mu^{(i)}$.
The definition of \citeauthor{mccann1997convexity} interpolation implies that these iterates all lie on the geodesic between $\mu$ and $\nu$.
This ideal progressive Monge problem precisely mimicks our progressive map estimator, though (1) these quantities are defined at the population level, and (2) the maps are defined a solutions to the Monge problem, rather than its entropic analogue. Recall that we write $\hat{\mu}$ and $\hat{\nu}$ as the empirical measures associated with $\mu$ and $\nu$, and recursively define $\myTent^{(i)}$ to be the entropic map from $\hat{\mu}^{(i)}_\eps$ to $\hat{\nu}$, where $\hat{\mu}^{(0)}_\eps \defeq \hat{\mu}$, and
\begin{align}
    \hat{\mu}_\eps^{(i+1)} \defeq (\myS^{(i)})_\# \hat{\mu}_\eps^{(i)} \defeq ((1-\alpha_i)\Id  + \alpha_i \myTent^{(i)})_\# \hat{\mu}_\eps^{(i)}\,.
\end{align}
We also require $\hat{R}_\eps^{(i)}$, defined to be the the entropic map between $\mu^{(i)}$ and $\hat{\nu}$ using regularization $\eps_i$. This map can also be seen as a ``one-sample" estimator, which starts from iterates of the McCann interpolation, and maps to an empirical target distribution.\looseness-1

To control the performance of $\hat{R}_\eps^{(i)}$ below, we want to use \cref{lem:one_sample}. To do so, we need to verify that $\mu^{(i)}$ also satisfies the key assumptions \textbf{(A1)} to \textbf{(A3)}. This is accomplished in the following lemma.

\begin{lemma}[Error rates for $\hat{R}_\eps^{(i)}$]\label{lem:hat_r_error}
For any $i \geq 0$, the measures $\mu^{(i)}$ and $\nu$ continue to satisfy \textbf{(A1)} to \textbf{(A3)}, and thus
\begin{align*}
    \E\|\hat{R}_\eps^{(i)} - T_0\|^2_{L^2(\mu^{(i)})} \lesssim \eps_i^{1-d/2}n^{-1/2} + \eps_i^{2}\,.
\end{align*}
\end{lemma}
\begin{proof}
We verify that the conditions \textbf{(A1)} to \textbf{(A3)} hold for the pair $(\mu^{(1)},\nu)$; repeating the argument for the other iterates is straightforward. 

First, we recall that for two measures $\mu_0 \defeq \mu,\mu_1 \defeq \nu$ with support in a convex subset $\Omega \subseteq \R^d$, the McCann interpolation $(\mu_\alpha)_{\alpha\in[0,1]}$ remains supported in $\Omega$; see \citet[Theorem 5.27]{santambrogio2015optimal}. Moreover, by Proposition 7.29 in \citet{santambrogio2015optimal}, it holds that
\begin{align*}
    \|\mu_\alpha\|_{L^\infty(\Omega)}\leq\max\{\|\mu_0\|_{L^\infty(\Omega)},\|\mu_1\|_{L^\infty(\Omega)}\} \leq \max\{\mu_{\max},\nu_{\max}\}\,,
\end{align*}
for any $\alpha\in[0,1]$, recall that the quantities $\mu_{\max},\nu_{\max}$ are from \textbf{(A1)}. Thus, the density of $\mu^{(1)} = \mu_{\alpha_0} = ((1-\alpha_0)\Id + \alpha_0 T)_\#\mu$ is uniformly upper bounded on $\Omega$; altogether this covers \textbf{(A1)}. For \textbf{(A2)} and \textbf{(A3)}, note that we are never leaving the geodesic. Rather than study the ``forward'' map, we can therefore instead consider the ``reverse" map 
\begin{align*}
    \bar{T} \defeq (\alpha_0\Id + (1-\alpha_0)T^{-1})\,,
\end{align*}
which satisfies ${\bar T}_\#\nu = \mu^{(1)}$ and hence $T^{(1)} =\bar{T}^{-1}$.
We now verify the requirements of \textbf{(A2)} and \textbf{(A3)}.
For \textbf{(A2)}, since $(T^{(1)})^{-1} = \bar T = \alpha_0\Id + (1-\alpha_0)T^{-1}$, and $T^{-1}$ is three-times continuously differentiable by assumption, the map $(T^{(1)})^{-1}$ is also three times continuously differentiable, with third derivative bounded by that of $T^{-1}$.
For \textbf{(A3)}, we use the fact that $T = \nabla \varphi_0$ for some function $\varphi_0$ which is $\Lambda$-smooth and $\lambda$-strongly convex.
Basic properties of convex conjugation then imply that $T^{(1)} = \nabla \varphi_1$, where
$\varphi_1 = (\alpha_0 \tfrac{\|\cdot\|^2}{2} + (1-\alpha_0) \varphi_0^*)^*$.
Since the conjugate of a $\lambda$-strongly convex function is $\lambda^{-1}$-smooth, and conversely, we obtain that the function $\varphi_1$ is $(\alpha_0 + (1-\alpha_0) \lambda^{-1})^{-1}$ strongly convex and
$(\alpha_0 + (1-\alpha_0) \Lambda^{-1})^{-1}$.
In particular, since $(\alpha_0 + (1-\alpha_0) \lambda^{-1})^{-1} \geq \min(1, \lambda)$ and $(\alpha_0 + (1-\alpha_0) \Lambda^{-1})^{-1} \leq \max(1, \Lambda)$, we obtain that $D T^{(1)}$ is uniformly bounded above and below.
\end{proof}

We define the following quantities which we will recursively bound:
\begin{align}
    \Delta_i \defeq \E\|\myTent^{(i)} - T^{(i)}\|^2_{L^2(\mu^{(i)})} \,, \Delta_{R_i} \defeq \|\hat{R}_\eps^{(i)} - T^{(i)}\|^2_{L^2(\mu^{(i)})}\,,
    \mathbb{W}_i \defeq \E W_2^2(\hat{\mu}_\eps^{(i)},\mu\ith{i})\,,
\end{align}
as well as
\begin{align*}
    \mathcal{A}_i \defeq 1 - \alpha_{i} + R^2 \frac{\alpha_{i}}{\eps_{i}}\,,
\end{align*}
where recall $R$ is the radius of the ball $B(0;R)$ in $\R^d$.

First, the following lemma:
\begin{lemma}\label{lem:summation_lemma}
If the support of $\mu$ and $\nu$ is contained in $B(0; R)$ and $\alpha_i \lesssim \eps_i$ for $i = 0, \dots, k$, then
\begin{align*}
\norm{\Tprog\ith{k} - T_0}^2_{L^2(\mu)}\lesssim  \sum_{i=0}^{k} \Delta_i\,.
\end{align*}
\end{lemma}
\begin{proof}
    We prove this lemma by iterating over the quantity defined by
\begin{align*}
    E_{j} \defeq \norm{\myTent\ith{k}\circ \myS\ith{k-1} \circ \dots \circ \myS\ith{j} -  T\ith{k}\circ  S\ith{k-1} \circ \dots  \circ S\ith{j} }^2_{L^2(\mu\ith{j})}.
\end{align*}
when $j\leq k-1$ and $E_k = \Delta_k$. By  adding and subtracting $S\ith{j}$ and $\hat{S}_\eps\ith{j}$ appropriately, we obtain for $j \leq k-1$,
\begin{align*}
    E_{j} & \lesssim  \norm{\myTent\ith{k}\circ \myS\ith{k-1} \circ \dots \circ \myS\ith{j} -  \myTent\ith{k}\circ \myS\ith{k-1} \circ \dots \circ \myS\ith{j+1} \circ S\ith{j}}^2_{L^2(\mu\ith{j})}\\
    &\qquad\qquad + \norm{\myTent\ith{k}\circ \myS\ith{k-1} \circ \dots \circ \myS\ith{j+1} \circ S\ith{j} -  T\ith{k}\circ  S^{(k-1)} \circ \dots  \circ S\ith{j} }^2_{L^2(\mu\ith{j})}\\
    & \leq \left(\alpha_{j}\mathrm{Lip}(\myTent\ith{k})\mathrm{Lip}(\myS\ith{k-1})\dots\mathrm{Lip}(\myS\ith{j+1})\right)^2\Delta_{j} + E_{j+1}\\
    & \lesssim \left( \frac{ \alpha_{j}}{\eps_k}\prod_{l=j+1}^{k-1}\mathcal{A}_\ell\right)^2\Delta_{j} + E_{j+1}\\
    & \lesssim \Delta_{j} + E_{j+1}
\end{align*}
where in the last inequality we have used the fact that $\alpha_j \lesssim \eps_j$, so that  $\mathcal{A}_k \lesssim 1$ for all $k$. Repeating this process yields $\norm{\Tprog\ith{k} - T_0}^2_{L^2(\mu)} = E_0 \lesssim \sum_{i=0}^k \Delta_k$, which completes the proof.
\end{proof}

To prove \cref{thm:consistency_multistep}, it therefore suffices to bound $\Delta_k$. We prove the following lemma by induction, which gives the proof.
\begin{lemma}\label{lem:recursion2}
Assume $d \geq 4$.
Suppose \textbf{(A1)} to \textbf{(A3)} hold, and $\alpha_i \asymp n^{-1/d}$ and $\eps_i \asymp n^{-1/2d}$ for all $i \in [k]$. Then it holds that for $k \geq 0$,
\begin{align*}
    \mathbb{W}_k \lesssim_{\log(n)} n^{-2/d}\,, \quad \text{and} \quad \Delta_k \lesssim_{\log(n)} n^{-1/d}\,.
\end{align*}
\end{lemma}
\begin{proof}
We proceed by induction. 
For the base case $k = 0$, the bounds of \citet{fournier2015rate} imply that $\mathbb{W}_0 = \E[W_2^2(\hat{\mu},\mu)]\lesssim n^{-2/d}$.
Similarly, by \cref{lem:two_sample}, we have $\Delta_0 \lesssim_{\log(n)} \eps_0^{-d/2} n^{-1/2} + \eps_0^2 \lesssim_{\log(n)} n^{-1/d}$.

Now, assume that the claimed bounds hold for $\mathbb{W}_k$ and $\Delta_k$.
We have
\begin{align*}
    \mathbb{W}_{k+1} & = \E[W_2^2((\myS\ith{k})_\# \hat\mu_\eps\ith{k},(S\ith{k})_\# \mu\ith{k})] \\
    &\lesssim \Exp\,\WD_2^2\left((\myS\ith{k})_\# \hat\mu_\eps\ith{k} , (\myS\ith{k})_\# \mu\ith{k}\right) +
        \Exp\,\WD_2^2\left((\myS\ith{k})_\#\mu\ith{k}, (S\ith{k})_\#\mu\ith{k}\right)\\
        & \leq \E [\mathrm{Lip}(\myS\ith{k})^2 W_2^2(\hat \mu_\eps\ith{k},\mu\ith{k})]
        + \E\norm{\myS\ith{k} - S\ith{k}}_{L^2(\mu\ith{k})}^2 \\
        & \lesssim {\cal A}_{k}^2\mathbb{W}_k + \alpha_{k}^2\Delta_k \\
        &\lesssim \mathbb{W}_k + \alpha_{k}^2\Delta_k \\
        & \lesssim n^{-2/d}\,,
\end{align*}
where the last step follows by the induction hypothesis and the choice of $\alpha_k$.
By \cref{prop:stability_phi} and the preceding bound, we have
\begin{align*}
    \Delta_{k+1} & \lesssim \|\myTent \ith{k+1} - \hat R_\eps\ith{k+1}\|^2_{L^2(\mu\ith{k+1})} + \Delta_{R_{k+1}} \\
    & \lesssim \eps_{k+1}^{-2} \mathbb{W}_{k+1} + \Delta_{R_{k+1}} \\
    & \lesssim \eps_{k+1}^{-2} n^{-2/d} + \Delta_{R_{k+1}}.
\end{align*}
\Cref{lem:hat_r_error} implies that $\Delta_{R_{k+1}} \lesssim_{\log n} n^{-1/d}$.
The choice of $\eps_{k+1}$ therefore implies $\Delta_{k+1} \lesssim_{\log(n)} n^{-1/d}$, completing the proof.
\end{proof}

\subsection{Proofs for the CIFAR Benchmark}
\begin{proof}[Proof of \cref{prop:cifar}]
    By \cref{prop:g_blur}, the Gaussian blur map $G$ is a linear positive-definite operator. Considering the \smash{$h = \tfrac{1}{2}\norm{\cdot}_2^2$} cost, then $G$ acts as a \citeauthor{Monge1781} map between from distribution $\hat \mu$ over a finite set of images, onto their blurred counterparts $\hat \nu = G_\#\hat \mu$. This is a direct corollary of \citeauthor{Bre91}'s Theorem, and follows the fact that $G$ is the gradient of $h(U) = \tfrac{1}{2}\langle U, G(U)\rangle$ the convex potential, and therefore a \citeauthor{Monge1781} map \citep[c.f. Section 1.3.1,][]{santambrogio2015optimal}.
Therefore, and again following \citeauthor{Bre91}, the optimal assignment between $\hat\mu$ and their blurred 
 counterparts $\hat\nu$, is necessarily that which maps an image to its blurred version regardless of the value of $\sigma<\infty$ and the optimal permutation is the identity.
\end{proof}

\begin{proposition}\label{prop:g_blur}
    The gaussian blur operator $G: U \rightarrow KUK$ is a linear positive-definite operator, where $U, K \in \sR^{N\times N}$ and the kernel $K$ is defined via
    \[K =\left[\exp\left(-(i-j)^2/(\sigma N^2)\right)\right]_{ij}, \quad \forall i,j \leq N.\]
\end{proposition}
\begin{proof}
The linearity of the operator is implied by the linearity of matrix multiplication.
As for the positive-definiteness, we show that the kernel matrix corresponding to the operator $G$ is positive-definite. 
Consider $s$ images $U_1, \dots, U_s$ in $\sR^{N\times N}$ and the corresponding kernel matrix $A \in \sR^{s\times s}$ defined as
\[
A_{ij} \coloneqq \langle U_i,\, G(U_j)\rangle = \langle U_i,\, KU_jK \rangle = \langle KU_i, \, KU_j\rangle.
\]
This is a dot-product matrix (of all elements $KU_i$), and is therefore always positive definite.
\end{proof}

\end{document}